%% file: paper.tex
\documentclass[anonymous=false, timestamp=false, review=false, acmtog]{acmart}

\AtBeginDocument{%
  \providecommand\BibTeX{{%
    \normalfont B\kern-0.5em{\scshape i\kern-0.25em b}\kern-0.8em\TeX}}}

\setcopyright{acmcopyright}
\copyrightyear{2020}
\acmYear{2020}
\usepackage{multirow}
\usepackage{cleveref}
\usepackage{siunitx}
\usepackage{booktabs}
\usepackage{adjustbox}
\usepackage{soul}

\begin{document}

%\title{Iterative Text-based Editing of Talking-head Video with Refinement and Performance Controls}
\title{Iterative Text-based Editing of Talking-heads Using Neural Retargeting}

\author{Xinwei Yao}
\email{xinwei.yao@cs.stanford.edu}
\affiliation{
  \institution{Stanford University}
  \department{Department of Computer Science}
}

\author{Ohad Fried}
\email{ohad.fried@post.idc.ac.il}
\affiliation{
  \institution{The Interdisciplinary Center Herzliya}
  \department{Department of Computer Science}
}

\author{Kayvon Fatahalian}
\email{kayvonf@cs.stanford.edu}
\affiliation{
  \institution{Stanford University}
  \department{Department of Computer Science}
}

\author{Maneesh Agrawala}
\email{maneesh@cs.stanford.edu}
\affiliation{
  \institution{Stanford University}
  \department{Department of Computer Science}
}

\renewcommand{\shortauthors}{Yao, et al.}

\input{abstract.tex}




\newcommand*{\ShowNotes}{}
\input{macros}

\begin{teaserfigure}
    \vspace{-1.em}
  \includegraphics[width=\textwidth]{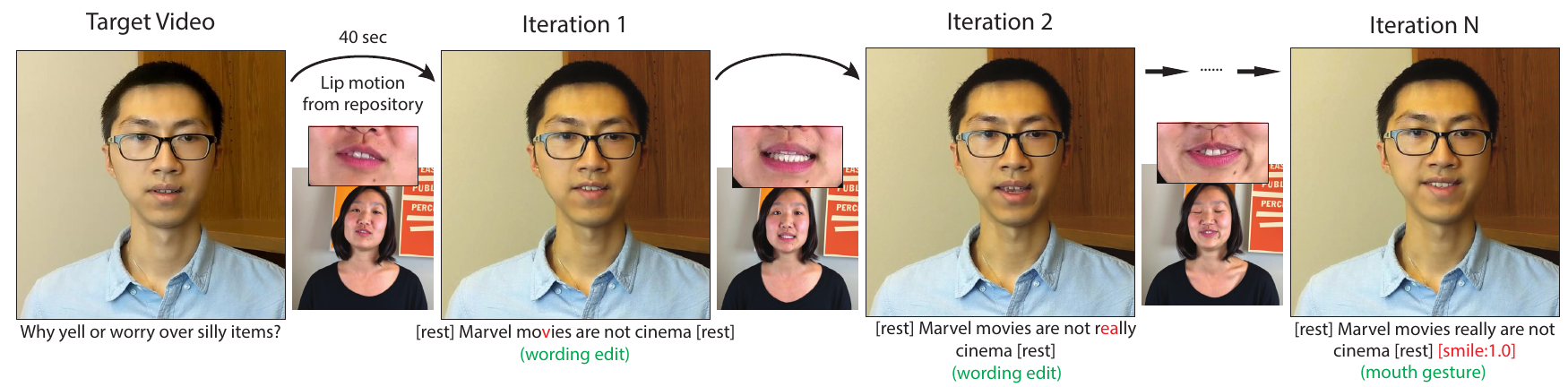}
  \vspace{-2.5em}
  \caption{
    Our iterative text-based tool for editing talking-head video takes
    2-3 minutes of a target video as input and is designed to support
    an iterative editing workflow. On each iteration the user might
    edit the wording of the speech (itr. 1 and 2), refine mouth
    motions if necessary to reduce artifacts, manipulate the performance by
    inserting mouth gestures (itr. N) or change the overall speaking style.
    Unlike previous techniques that require hours
    our tool takes about 40 seconds to generate each iteration,
    making it practical for users to explore a variety of different
    edits as they iterate. 
    Our approach is to retarget lip motion from a repository of source
    actor video to the target actor. The frame shown for each iteration
    corresponds to the red edit text/gesture below the frame.
    %% We propose a system that can iteratively edit talking head
    %% video from text, and here we present one example editing session
    %% with the system, where we explore different ways to say that
    %% Marvel movies should not be considered as cinema.  We first try
    %% the line ``Marvel movies aren't cinema'', and the system produces
    %% a video of the target actor saying exactly this line with lip
    %% motions synchronized to the audio with a synthetic voice.  Feeling
    %% it is too blunt and may cause a controversy, we slightly change
    %% the wording to ``Marvel movies aren't really cinema'' in the next
    %% iteration, inspect the result from the system again and keep
    %% iterating.  During the iteration, we can correct errors from the
    %% automatic synthesis pipeline at any point with refinement controls
    %% such as smoothing out a particular jumpy transition.  After a few
    %% iterations we eventually settle on the firmer statement ``Marvel
    %% movies really aren't cinema'', and decide to add a smile of length
    %% 0.8 seconds at the beginning of the sentence to soften the overall
    %% tone.  The system produces the video with the desired mouth
    %% gesture and spoken content with a seamless audio-visual flow.
    %% Later if the user obtain a real recording of the target actor
    %% saying the line, they can re-run the synthesis pipeline with
    %% timings from the real audio.
  }
  \label{fig:teaser}
\end{teaserfigure}

\maketitle

\input{intro}
\input{relatedWork}
\input{method}
\input{results}

\input{evaluation}

\input{futureWork}

\input{ethics}

\input{conclusion}

% \begin{acks}
% \end{acks}

\bibliographystyle{ACM-Reference-Format}
\bibliography{references}

\appendix
\end{document}

\title{Iterative Text-based Editing of Talking-heads Using Neural Retargeting: Supplementary Material}

\author{Xinwei Yao}
\email{xinwei.yao@cs.stanford.edu}
\affiliation{
  \institution{Stanford University}
  \department{Department of Computer Science}
}

\author{Ohad Fried}
\email{ohad.fried@post.idc.ac.il}
\affiliation{
  \institution{The Interdisciplinary Center Herzliya}
  \department{Department of Computer Science}
}

\author{Kayvon Fatahalian}
\email{kayvonf@cs.stanford.edu}
\affiliation{
  \institution{Stanford University}
  \department{Department of Computer Science}
}

\author{Maneesh Agrawala}
\email{maneesh@cs.stanford.edu}
\affiliation{
  \institution{Stanford University}
  \department{Department of Computer Science}
}

\renewcommand{\shortauthors}{Yao, et al.}

\newcommand*{\ShowNotes}{}
\input{macros}

\maketitle

\section{Adjusted P-Values for User Studies}
\new{
We apply Tukey's range test for pair-wise comparisons, and report the p-values of each pair (adjusted for multiple testing) in \Cref{tab:pvalues}.
}

\section{Synthetic Voice User Study}
Using the same design as our other user studies, our synthetic voice user study
aims to rule out the hypothesis that using a
synthesized voice in our tool is already so unrealistic, that there is
no point in employing our sophisticated video synthesis technique.
We recruited 73 participants to view 8 videos each, and compare a
baseline method that linearly retimes the target video to match the
length of the speech given by the edit, to results from an early version of our tool
both automatically generated and with manual refinement (\Cref{tab:user_study_synth}).
The difference between conditions is statistically significant
(Kruskal-Wallis test, $p < 10^{-9}$). Both our automatic synthesis
pipeline and our pipeline with manual refinement perform better than
the baseline algorithm (Tukey's range test, $p = 0.02$ 
%for automatic pipeline and $p = 0.001$ for manually refined results). 
and $p = 0.001$ respectively).
These results
suggest that synthesizing video to match audio increases realism, even
when using synthesized speech audio.

\begin{table}
\footnotesize
\centering
\begin{tabular}{@{}lllcc@{}}
\toprule
& Group 1 & Group 2 & Mean Difference & Adjusted p-value \\
\midrule
\parbox[t]{0mm}{\multirow{10}{*}{\rotatebox[origin=c]{90}{\footnotesize Short Phrase}}}
& Fried [2019] (> 1 hr)   & Fried [2019] (< 5 min)   & -0.3127  & \textbf{0.0065} \\
& Fried [2019] (> 1 hr)   & Modified Fried (> 1 hr)  & 0.319    & \textbf{0.0051} \\
& Fried [2019] (> 1 hr)   & Ground-truth             & 0.6798   & \textbf{0.001}  \\
& Fried [2019] (> 1 hr)   & Ours (< 5 min)           & 0.3067   & \textbf{0.0072} \\
& Fried [2019] (< 5 min)  & Modified Fried (> 1 hr)  & 0.6317   & \textbf{0.001}  \\
& Fried [2019] (< 5 min)  & Ground-truth             & 0.9926   & \textbf{0.001}  \\
& Fried [2019] (< 5 min)  & Ours (< 5 min)           & 0.6194   & \textbf{0.001}  \\
& Modified Fried (> 1 hr) & Ground-truth             & 0.3609   & \textbf{0.0011} \\
& Modified Fried (> 1 hr) & Ours (< 5 min)           & -0.0123  & 0.9    \\
& Ground-truth            & Ours (< 5 min)           & -0.3732  & \textbf{0.001}  \\
\midrule
\parbox[t]{0mm}{\multirow{10}{*}{\rotatebox[origin=c]{90}{\footnotesize Full Sentence}}}
& Fried [2019] (> 1 hr)   & Fried [2019] (< 5 min)   & -0.0185  & 0.9 \\
& Fried [2019] (> 1 hr)   & Modified Fried (> 1 hr)  & 0.3303   & \textbf{0.001} \\
& Fried [2019] (> 1 hr)   & Ground-truth             & 1.2914   & \textbf{0.001}  \\
& Fried [2019] (> 1 hr)   & Ours (< 5 min)           & 0.6224   & \textbf{0.001} \\
& Fried [2019] (< 5 min)  & Modified Fried (> 1 hr)  & 0.3488   & \textbf{0.001}  \\
& Fried [2019] (< 5 min)  & Ground-truth             & 1.3098   & \textbf{0.001}  \\
& Fried [2019] (< 5 min)  & Ours (< 5 min)           & 0.6408   & \textbf{0.001}  \\
& Modified Fried (> 1 hr) & Ground-truth             & 0.961    & \textbf{0.001} \\
& Modified Fried (> 1 hr) & Ours (< 5 min)           & 0.292    & \textbf{0.0058}    \\
& Ground-truth            & Ours (< 5 min)           & -0.669   & \textbf{0.001}  \\
\bottomrule
\end{tabular}
\caption{
\new{
  Adjusted p-values for all pair-wise comparisons in user studies.
  ``Mean Difference'' is mean score of Group 2 minus that of Group 1.
  Significant pair-wise comparisons (p<=0.05) have bolded p-values.
}
}
\label{tab:pvalues}
\end{table}

\begin{table}
\footnotesize
\setlength{\tabcolsep}{7.5pt}
\centering
\begin{tabular}{@{}lccccccc@{}}
\toprule
          & \multicolumn{5}{c}{Likert response (\%)} & & \\
\cmidrule(lr){2-6}
Condition & 5 & 4 & 3 & 2 & 1 & Mean  & `Real' \\
%                     & \rot{(5) Strongly agree}           & \rot{(4) Agree}      & \rot{(3) Neither agree nor disagree}      & \rot{(2) Disagree}      & \rot{(1) Strongly disagree}             & Mean  & \% `Real' \\
%                     & \hspace{0.1cm}agree)  &        &        &        & \hspace{0.44cm}disagree) & score &           \\
\midrule

Baseline             & 3.5                 & 12.3 &  7.3 & 30.1 & 46.8                   & 2.0   & 15.8\%    \\
Ours (automatic)     & 4.1                 & 19.2 & 16.4 & 31.5 & 28.8                   & 2.4   & 23.3\%    \\
Ours (refined)       & 5.8                 & 27.0 & 13.3 & 30.3 & 23.6                   & 2.6   & 32.8\%    \\
\bottomrule
\end{tabular}
\caption{User study results for videos with synthesized audio. We compare a video retiming baseline with our method with and without manual refinement. We report percentage of each answer on a 5-Point Likert scale, as well as mean score and percent of video that received a score of 4 or 5 (`real'). 
Both our pipelines outperform the baseline algorithm, which suggests that synthesizing video to match audio increases realism, even when using synthesized speech audio.
}
\label{tab:user_study_synth}
\end{table}

\bibliographystyle{ACM-Reference-Format}
\bibliography{references}

\appendix

%% file: abstract.tex
\begin{abstract}
%\maneesh{Alternate titles:\\ Prototyping Dialogue via Interactive
%Text-based Editing of Talking-head Video \\ Protyping Talking-head
%Dialogue via Interactive Video ReWrite\\Interactive Editing of
%Talking-Head Dialogue and Performance\\Interactive Editing of
%Talking-Head Video with Performance Controls}

%\ohad{I much prefer not to have ``prototyping'' in the title.}
%\ohad{We do not say anything about the fact that the editing is
%text-based. It might be fine, but on the other hand something like
%``Interactive Editing of Talking-Head Video with Performance
%Controls'' can just as well be the Adobe Premiere interface. One way
%to resolve this is to add a project name. Something like: ``Text2Vid:
  %Interactive Editing of Talking-head Dialogue and Performance''}
  %\maneesh{Too similar to fig 1 caption?}
%\david{Abstract word limit for ToG is 200 words. Need to shorten more.}
We present a text-based tool for editing talking-head video
that enables an iterative editing workflow.
On each iteration users can edit the wording of the speech, further
refine mouth motions if necessary to reduce artifacts and manipulate
non-verbal aspects of the performance by inserting mouth
gestures (e.g. a smile) or changing the overall performance style
(e.g. energetic, mumble).
Our tool requires only 2-3 minutes of the target actor video and
it synthesizes the video for each iteration in about 40 seconds, 
allowing users to quickly explore many editing possibilities as they iterate.
Our approach is based on two key ideas.
(1)
We develop a fast phoneme search algorithm that can
quickly identify phoneme-level subsequences of the source repository
video that best match a desired edit. This enables our fast iteration
loop.
% \david{We develop a fast phoneme search algorithm that can quickly
%   identify and stitch together phoneme-level subsequences of source
%   repository video to best match a desired edit.} \maneesh{Not sure
%   how prev. sentence differs from the one we used previously. I kind
%   of like the orig version better -- see next sentence.}
(2)
We leverage a large repository of video of a source actor and
develop a new self-supervised neural retargeting technique for
transferring the mouth motions of the source actor to the target
actor.  This allows us to work with relatively short target actor
videos, making our approach applicable in many real-world editing
scenarios. 
Finally, our refinement and performance controls give users the
ability to further fine-tune the synthesized results.
\end{abstract}

% LocalWords:  retargeting subsequences

%% file: macros.tex
%%%%% For comments:
\newcommand{\ignorethis}[1]{}
\newcommand{\redund}[1]{#1}

%%%%% Latin and language:
% \newcommand{\etal       }     {\textit{et~al.}} old; not like ACM style
\newcommand{\etal       }     {{et~al.}}
\newcommand{\apriori    }     {\textit{a~priori}}
\newcommand{\aposteriori}     {\textit{a~posteriori}}
\newcommand{\perse      }     {\textit{per~se}}
\newcommand{\eg         }     {{e.g.~}}
\newcommand{\Eg         }     {{E.g.~}}
\newcommand{\ie         }     {{i.e.~}}
\newcommand{\naive      }     {{na\"{\i}ve}}

%%%%% Math symbols:
\newcommand{\Identity   }     {\mat{I}}
\newcommand{\Zero       }     {\mathbf{0}}
\newcommand{\Reals      }     {{\textrm{I\kern-0.18em R}}}
\newcommand{\isdefined  }     {\mbox{\hspace{0.5ex}:=\hspace{0.5ex}}}
\newcommand{\texthalf   }     {\ensuremath{\textstyle\frac{1}{2}}}
\newcommand{\half       }     {\ensuremath{\frac{1}{2}}}
\newcommand{\third      }     {\ensuremath{\frac{1}{3}}}
\newcommand{\fourth     }     {\ensuremath{\frac{1}{4}}}

%\newcommand{\degree} {\ensuremath{^{\circ}}}

%%%%% Math modifiers:
\newcommand{\mat        } [1] {{\text{\boldmath $\mathbit{#1}$}}}
\newcommand{\Approx     } [1] {\widetilde{#1}}
\newcommand{\change     } [1] {\mbox{{\footnotesize $\Delta$} \kern-3pt}#1}

%%%%% Math functions:
\newcommand{\Order      } [1] {O(#1)}
\newcommand{\set        } [1] {{\lbrace #1 \rbrace}}
\newcommand{\floor      } [1] {{\lfloor #1 \rfloor}}
\newcommand{\ceil       } [1] {{\lceil  #1 \rceil }}
\newcommand{\inverse    } [1] {{#1}^{-1}}
\newcommand{\transpose  } [1] {{#1}^\mathrm{T}}
\newcommand{\invtransp  } [1] {{#1}^{-\mathrm{T}}}
\newcommand{\relu       } [1] {{\lbrack #1 \rbrack_+}}

%%%%% Math functions with small (fixed) and large (expandable) forms:
\newcommand{\abs        } [1] {{| #1 |}}
\newcommand{\Abs        } [1] {{\left| #1 \right|}}
\newcommand{\norm       } [1] {{\| #1 \|}}
\newcommand{\Norm       } [1] {{\left\| #1 \right\|}}
\newcommand{\pnorm      } [2] {\norm{#1}_{#2}}
\newcommand{\Pnorm      } [2] {\Norm{#1}_{#2}}
\newcommand{\inner      } [2] {{\langle {#1} \, | \, {#2} \rangle}}
\newcommand{\Inner      } [2] {{\left\langle \begin{array}{@{}c|c@{}}
                               \displaystyle {#1} & \displaystyle {#2}
                               \end{array} \right\rangle}}

%%%%% A two part math definition
\newcommand{\twopartdef}[4]
{
  \left\{
  \begin{array}{ll}
    #1 & \mbox{if } #2 \\
    #3 & \mbox{if } #4
  \end{array}
  \right.
}

%%%%% A four part math definition
\newcommand{\fourpartdef}[8]
{
  \left\{
  \begin{array}{ll}
    #1 & \mbox{if } #2 \\
    #3 & \mbox{if } #4 \\
    #5 & \mbox{if } #6 \\
    #7 & \mbox{if } #8
  \end{array}
  \right.
}

% Length of something
\newcommand{\len}[1]{\text{len}(#1)}

%%%%% Paper-specific stuff:

% reduce hyphenation (slay the hyper hyphenator with 2000)
%\pretolerance 800

% These variables are for width and height and gaps in figures:
% set with something like: \setlength{\h}{1cm}
\newlength{\w}
\newlength{\h}
\newlength{\x}

% maybe requires \usepackage[usenames]{color}
\definecolor{darkred}{rgb}{0.7,0.1,0.1}
%\definecolor{darkgreen}{rgb}{0.1,0.5,0.1}
\definecolor{darkgreen}{rgb}{0.1,0.8,0.1}
\definecolor{cyan}{rgb}{0.7,0.0,0.7}
\definecolor{otherblue}{rgb}{0.1,0.4,0.8}
\definecolor{maroon}{rgb}{0.76,.13,.28}
\definecolor{burntorange}{rgb}{0.81,.33,0}

% see line at top of main file to show/hide notes
\ifdefined\ShowNotes
  \newcommand{\colornote}[3]{{\color{#1}\bf{#2 #3}\normalfont}}
\else
  \newcommand{\colornote}[3]{}
\fi

\newcommand {\note}[1]{\colornote{maroon}{}{#1}}
\newcommand {\todo}[1]{\colornote{cyan}{TODO}{#1}}
\newcommand {\david}[1]{\colornote{blue}{DY:}{#1}}
\newcommand {\ohad}[1]{\colornote{darkgreen}{OF:}{#1}}
\newcommand {\michael}[1]{\colornote{cyan}{MZ:}{#1}}
\newcommand {\kayvon}[1]{\colornote{otherblue}{KF:}{#1}}
\newcommand {\maneesh}[1]{\colornote{red}{MA:}{#1}}

\newcommand {\new}[1]{#1}
% \newcommand {\new}[1]{{\color{blue}{#1}\normalfont}}

%\newcommand*\rot[1]{\rotatebox{90}{#1}}

% To rotate within a table:
\newcolumntype{R}[2]{%
    >{\adjustbox{angle=#1,lap=\width-(#2)}\bgroup}%
    l%
    <{\egroup}%
}
\newcommand*\rot{\multicolumn{1}{R{45}{1em}}}

\newcommand {\newstuff}[1]{#1}

\newcommand\todosilent[1]{}

% For checkmark and x mark
% Might require \usepackage{pifont}
% \newcommand{\cmark}{\textcolor{darkgreen}{\ding{51}}}%
% \newcommand{\xmark}{\textcolor{darkred}{\ding{55}}}%

%\ifdefined\SmallImages
%  \newcommand{\images}{{images-small}}
%\else
%  \newcommand{\images}{{images}}
%\fi

% To prevent shortcite error (uncomment if shortcite does not exist)
%\newcommand{\shortcite}[1]{\cite{#1}}

% New commands for algorithms
%\renewcommand{\algorithmicrequire}{\textbf{Input:}}
%\renewcommand{\algorithmicforall}{\textbf{for each}}

%\newcommand{\projurl}{http://www.abcde.com/}
%\newcommand{\projectpage}{\href{\projurl}{\small\textbf{\texttt{\projurl}}}}

% add support for multiple markers referencing the same footnote
%\makeatletter
%\newcommand\footnoteref[1]{\protected@xdef\@thefnmark{\ref{#1}}\@footnotemark}
%\makeatother

% Use the following to display page and column width
% Points:
%\the\columnwidth
% Any units:
%Column width \printlen[3][cm]{\columnwidth} \\
%Text width \printlen[3][cm]{\textwidth} \\

% New commands for algorithms
% \newcommand{\alglinenum } [1] {\ref{#1}}
% \newcommand{\algline    } [1] {Line~\alglinenum{#1}}
% \renewcommand{\algorithmicrequire}{\textbf{Input:}}
% \renewcommand{\algorithmicforall}{\textbf{for each}}

\newcommand{\cmatch}{C_{\mbox{match}}}

%% file: intro.tex
\section{Introduction}

Tools for editing talking-head video using transcripts
have made it possible to easily remove filler words, emphasize phrases,
correct mistakes, and try different wordings of the
speech\,\cite{berthouzoz12,Fried_2019,thies2019neural,10.1145/3072959.3073640}.
%
%% Other tools use a new audio recording to puppeteer a talking-head
%% \cite{bregler97,10.1145/3072959.3073640,thies2019neural}. Such tools
%% can be extended via a text-to-speech system to allow transcript based
%% editing \cite{thies2019neural} or to summarize recordings
%% \cite{10.1145/3072959.3073640}.
%%
%\Ohad{I tried to make the statement more accurate by splitting into two groups. This way we can also talk about a few more audio based methods if we want to.}
% \ohad{Bregler and Obama are audio based. Neural Voice Puppetry is mostly audio based but also mentions text. I would remove Bregler and Obama. ``Time aligned'' does not apply to NVP, and also it does not require lots of data which clashes with the rest of the intro; I would remove NVP (and discuss it later on).} 
Many of these tools can synthesize high-quality results that
closely match the appearance of the unedited video.
Such tools have the potential to enable a variety of
post-capture editing applications including re-phrasing dialogue in a
film scene, dubbing commercials to a new language, developing
dialogue for a conversational video assistant, and fixing wording
errors in an online lecture.

Yet, current transcript-based video editing tools are impractical
for use in many real-world editing scenarios for four main reasons.
% \maneesh{May want to rearrange put slow feedback first -- not sure.}
%\david{Put slow feedback first to be consistent with video presentation.}

\vspace{0.5em}
\noindent
{\bf \em (1) Slow feedback loop hinders iterative editing.}
Synthesizing the edited result at high-quality is often extremely
slow.  For example, while viewers report that Fried et
al.'s\,\shortcite{Fried_2019} results appear very realistic, their approach
takes hours to generate a few seconds of edited video.  The slow
feedback loop --- time between specifying an edit and seeing the
result --- significantly hinders iterative editing (e.g. trying
different phrasings of dialogue).

\vspace{0.5em}
\noindent
{\bf \em (2) Require hours of target talking-head video.}  To
produce realistic results, many of these
tools\,\cite{Fried_2019,10.1145/3072959.3073640} require
hours of video of the target talking-head
actor. Some tools further require the actor to speak a set of
specialized phrases (e.g. TIMIT corpus\,\cite{garofolo1993timit}).
%\maneesh{Also is there a
%  requirement that the actor be captured in the same environment with
%  a static camera?  Maybe should reference the ones that have this
%  requirement. We should at least know which of them have this
%  requirement.}
%\ohad{We need to compose a list of all the papers
%    we want to mention in the intro. If we want more than a couple, it
%    will have to include audio based and perhaps video2video stuff,
% which means we need to tweak the first paragraph a bit.}
  In practice however, many video editing projects lack access to such
  large amounts of target actor video.

\vspace{0.5em}  
\noindent
{\bf \em (3) Missing controls for refining results.}
  None of these editing tools provide controls for manually refining the lip
  motions of the synthesized results, making it impossible to fix
  objectionable artifacts these editing tools sometimes generate (e.g. mouth
  doesn't fully close on \textbackslash m, \textbackslash b,
  \textbackslash p phonemes). 

\vspace{0.5em}  
\noindent
    {\bf \em (4) Missing controls for adjusting non-verbal
      performance.}  None of these editing tools include controls for
    adjusting the target actor's non-verbal performance by inserting
    mouth gestures (e.g. a smile) or changing the overall speaking
    style (e.g. mumbling, energetic).
    %The lack of such performance
    %controls limits how much users can manipulate the video.
  
\vspace{0.5em}
In this work we present an iterative talking-head video
editing tool that explicitly addresses all four of these issues.
While our approach builds on the high-quality synthesis technique of
Fried~et~al.\,\shortcite{Fried_2019}, we make several new
contributions.  We significantly reduce the time required to
synthesize video (from hours to about 40 seconds for a 6 word edit) by
developing a fast algorithm for searching the source repository for
the desired lip motions.  We lower the data requirement on the target
actor video (2-3 minutes are usually enough) by leveraging a large
repository of video from a source actor and use a new self-supervised
neural retargeting technique to transfer their lip motions to the
target actor.  We provide controls to refine results by allowing users
to smooth over jumpy transitions and force mouth closure on the
results of the automated synthesis pipeline. Finally, we enable
insertion of non-verbal mouth gestures with the same text interface,
as well as controls to switch between different speaking styles by
using a version of the source repository with the desired style.
%\maneesh{Perhaps we should be consistent and just call our
%retargeting ``neural retargeting'' everywhere. Perhaps call it
%self-supervised neural retargeting.}

As shown in Figure~\ref{fig:teaser} our tool enables an iterative
editing workflow. Given a short video of the target actor, the user
can edit the transcript and our tool synthesizes the corresponding
video in under a minute. 
The user can inspect the feedback and further adjust wording, refine the
lip motions and/or insert mouth gestures and quickly see how the
adjustment affects the synthesized video.
%
%\maneesh{Maybe move note to very end of intro?}
Note that our work focuses on generating video from text;
to obtain the corresponding speech audio, we rely on either having
access to the actor speaking the new content (e.g. from a prerecorded library of the actor's speech or recorded by the actor in real-time during editing), 
text-to-speech voice synthesis\,\cite{wavenet2016} or voice
cloning\,\cite{kumar2019lyrebird, jia2018sv2tts}. 
%Once a transcript has been finalized, the user may choose to
%re-capture an audio recording of the target actor saying the finalized
%lines and use our tool to generate the video to match the actor's
%final speech.
%\david{Maybe talk about the real-voice iteration mode here.} \maneesh{We probably should say this and combine somehow with previous sentence. Probably should also cite lyrebird for tts or voice cloning.}
%\ohad{I would kill final sentence and instead of a whole paragraph about results and eval, try to make that our concluding sentence. Something like "We demonstrate various editing scenarios using our tool, and evaluate result quality in two user studies"}
%Our videos include many results with a synthetic
%voice as well as the target actor's real voice for the finalized lines.
%\maneesh{Kill prev sentence?}
%

We demonstrate a variety of iterative editing sessions facilitated by
our tool and we conduct user studies which show that our synthesized
results are rated as ``real'' for 56.2\% of the sentence-long edits
and for 64.9\% of the phrase-long edits -- slightly better than the
previous state-of-the-art approach of Fried~et~al.\,\shortcite{Fried_2019}.
%\maneesh{User studies really
%  only compare to Fried19 and are inconclusive w.r.t NVP, so I changes prev. sentence. Though it is awkward to talk about Fried19 as previous state-of-the-art. -- could call it ``high-quality approach''}
Together
these results suggest that our algorithm provides the speed, data
efficiency and controls necessary for a practical iterative editing
workflow while maintaining high-quality synthesis results.

%% file: relatedWork.tex
\section{Related Work}

%% Our work builds on previous efforts in various sub-domains, including
%% neural rendering, dubbing, video to video translation, text analysis
%% and audio analysis. In the following we give details on methods which
%% are most closely related to ours --- those that aim to edit and
%% synthesize speaking humans.

%% Our work is most closely related to previous efforts on editing and
%% synthesizing speaking humans. \maneesh{kill for space?}  \ohad{I
%% prefer the slightly longer version that we had before. There are
%% many related bodies of work that we aren't even mentioning, so I
%% want readers to realize it was a deliberate choice.}

% ---------------------------------------------------------------------------

\paragraph{Video-driven talking-head synthesis.}
A common approach to synthesizing a talking-head video is to use
a ``driving'' video from a different actor
that has the desired
motion, expression and speech, and transfer those elements to the
primary talking-head.
%This reframes the synthesis as a decomposition and domain transfer
%problem --- trying to extract motion information from the driving
%video, and somehow transfer it onto the output.
Early attempts used facial landmarks from a video to retrieve frames
of a different person and play them back directly\,\cite{KemelSSS2010}
or after warping\,\cite{GarriVRTPT2014}.
% Early attempts used facial landmarks from a video to retrieve and play
% back frames of a different person\,\cite{KemelSSS2010}.  While these
% methods demonstrate the power of matching head pose and expression,
% their results resemble a stop-motion sequence and not a smooth
% video. Adding warping and blending allows the creation of face-swap
% videos\,\cite{GarriVRTPT2014}.
% \maneesh{To shorten, can we replace two previous sentences with ....
% Early attempts used facial landmarks from a video to retrieve and play
% back frames of a different person\,\cite{KemelSSS2010,GarriVRTPT2014}.}
% \ohad{If we do this, I would replace ``retrieve and play
% back frames'' with something that also applies to \cite{GarriVRTPT2014}}
%While these
%methods demonstrate the power of matching head pose and expression,
%their results resemble a stop-motion sequence and not a smooth
%video. Adding warping and blending allows the creation of face-swap
%videos\,\cite{GarriVRTPT2014}.
%
Opting for a lower data requirement, several approaches synthesize
video given only one or a few photos of the target person, either by
morphing and blending\,\cite{AverbCKC2017} or using neural
networks\,\cite{Geng:2018,Wiles18,zakharov2019fewshot,Pumarola_ijcv2019}. These
methods are successful in producing short expression videos, but are less
convincing for full sentences.
%\maneesh{Can we describe what goes wrong in a couple of words at the end of the previous sentence?}
%In this work we use short target videos instead of single frames,
%which allows us to produce higher quality results.
Several approaches use a tracked head model,
to decouple properties (e.g. pose, identity, expressions) to
produce convincing results\,\cite{VlasiBPP2005,GarriVSSVPT2015,ThiesZSTN2016a,kim2018DeepVideo,Kim19NeuralDubbing}.
We similarly
use a tracked head model to decouple such properties.
%Others have
%identified, like us, that transferring such parameters across people
%is not straightforward and requires special care
%\cite{Kim19NeuralDubbing}.
%\maneesh{Not sure what we are getting at
%  with prev. sentence. It seems unnecessary.}  \david{It shows that
%  some before us have found that retargeting is tricky and have
%  proposed their solutions.}
%
%Dynamic avatars can be rendered in
%realtime\,\cite{10.1145/2897824.2925873}, even on a mobile
%phone\,\cite{Nagano:2018}. \maneesh{Not sure how dynamic avatars
%  relates to others in this section.  seems out of place. Can we
%  kill?}
%
%Notably, Kim et al.\,\shortcite{Kim19NeuralDubbing} develop an
%unsupervised neural retargeting approach based on CycleGAN\,\cite{} to
%map parameters from one tracked model to another. In contrast we
%develop a self-supervised neural retargeting method.
All of these previous methods require a driving video to
specify the desired output head motion and expression. In this work we
specify those properties via text, which is often a simpler, lower-cost
interaction.

%Deep video portraits \cite{kim2018DeepVideo}

% Bringing Portraits to Life \cite{AverbCKC2017} (single photo to short expression video, not learning based)

% Warp-guided GANs for single-photo facial animation \cite{Geng:2018} (single photo to short expression video, GANs)

%Being John Malkovich \cite{KemelSSS2010} (not learning based, uses frame matching as is without blending)

%Automatic Face Reenactment \cite{GarriVRTPT2014} (not learning based, uses frame matching + face swap)

%Face2Face: Real-time Face Capture and Reenactment of RGB Videos \cite{ThiesZSTN2016a} (not learning; expression transfer + frame matching + mouth swap)

%Face Transfer with Multilinear Models \cite{VlasiBPP2005} (not learning; early attempt to decouple expression from speech visemes)

% X2Face: A network for controlling face generation using images, audio, and pose codes \cite{Wiles18} (synthesis from one or few frames, using driving video or audio)

%\cite{GarriVSSVPT2015}

% Video avatars
%paGAN: Real-time Avatars Using Dynamic Textures \cite{Nagano:2018} (controllable in realtime, after building an avatar from a single image)
%Real-time Facial Animation with Image-based Dynamic Avatars \cite{10.1145/2897824.2925873} (non learning; capture 32 images of someone, fit model and use warp+blend to synthesize) 

% add the few shot paper and the talking mona lisa paper (are they the same?) to the "from one or few examples" section

% ---------------------------------------------------------------------------

\paragraph{Voice-driven talking-head synthesis.}
Another approach for talking-head synthesis is to drive it with
voice. The pioneering work of Bregler et al.\,\shortcite{bregler97}
creates talking-heads through a combination of alignment and blending,
and was improved upon in various
followups\,\cite{EzzatGP2002,ChangE2005,LiuO2011}.
Others have used human voice-driven synthesis to dub
non-humans\,\cite{.20191220}.  Several methods synthesize a
talking-head given only one or a few video frames in addition to the
voice
track\,\cite{Chung17b,vougioukas2018endtoend,chen2019hierarchical,zhou2018talking,ijcai2019-129,Vougioukas2019}.
%\maneesh{How do prev. references differ from the ones mentioned in
%the previous paragraph that are driven by a few images --- do they
%have a voice track?}
However, the result is a fixed frame with a moving inner-face region,
or a tightly cropped head, and is easily distinguishable from
realistic video.  Suwajanakorn et
al.\,\shortcite{10.1145/3072959.3073640} demonstrate that using a
large repository of video (17 hours) can produce convincing synthesis
results.  In work concurrent to ours, Thies et
al.\,\shortcite{thies2019neural} produce video from speech. We compare
to their results in \Cref{subsec:compare_others}.
%\maneesh{We never in
%  this paper talk about the speed or data requirements of Thies19. I'm
%  not sure if we should. I also can't tell exactly what the right
%  comparison is in terms of speed. We should mention that Thies
%  doesn't have refinement and gesture controls somewhere -- I don't
%  think it really could.}
None of these voice-drive methods provide refinement and performance
controls that are essential for iterative editing.

\paragraph{Text-driven talking-head synthesis.}

Most related to our work are methods that perform text-based video
editing and synthesis.
%
%% Pavel et
%% al.\,\shortcite{pavel2014,pavel2016} incorporate a text interface to
%% summerize lecture videos and for video review and annotation.
%% %generate structured summaries of lecture videos [Pavel et al. 2014] \cite{pavel2014}
%% %annotate video with review feedback [Pavel et al.2016] \cite{pavel2016}
%% %
%% Leake et al.\,\shortcite{Leake:2017:CVE:3072959.3073653} 
%% %use the structure imposed by time-aligned transcripts to 
%% automatically edit together multiple takes of a scripted scene based on time-aligned transcripts and higher-level cinematic idioms.
%% %specified by the editor.
%% %
%% \ohad{If we want to keep this sub-section short, we can skip the Pavel
%%   and Leake citations. Maneesh --- this removes 3 citations of your
%%   work so I'll let you decide.}
%% %
Wang et al.\,\shortcite{wang2011text-driven} synthesize a
talking-head, and allow control over facial expressions, but the head
is floating in space and is not part of a photorealistic video.
Mattheyses et al.\,\shortcite{DBLP:conf/avsp/MattheysesLV10}
synthesize audio-visual speech from text, but the resulting videos have no
head motion making them unrealistic.
Berthouzoz et al.\,\shortcite{berthouzoz12} can edit
talking-head video by cutting, copying and pasting transcript
text. However, they do not allow synthesis of new words to change
phrasing or fix flubbed lines.
ObamaNet\,\cite{kumar2017obamanet} synthesizes both audio and video from text, using a large dataset of 17 hours of the president's speeches. 
While predominantly an audio-based method, Thies et al.\,\shortcite{thies2019neural} also show text-based results by incorporating a text-to-speech system. 
The work of Fried et al.\,\shortcite{Fried_2019} most closely
resembles ours, but it requires over 1 hour of target video and takes
hours to produce a result, while our tool requires 2--3 minutes of
target video and produces results in about 40 seconds.  We compare our results to
both Thies et al. and Fried et al. in Sections~\ref{subsec:compare_others}~and~\ref{sec:user_study} and find that the quality of our results is similar to both of these techniques.
%
%\maneesh{Perhaps mention that our quality is similar to both of these techniques, but unlike Thies we provide controls and unlike Fried we require less data and are much much faster.}
%Our result quality is similar to both of these techniques, and
Moreover we provide refinement and performance controls that are missing in all previous text-driven talking-head synthesis tools, but are critical for a practical video editing tool.

%% file: method.tex
\begin{figure}
  \centering
    \vspace{-1em}
  \includegraphics[width=\linewidth]{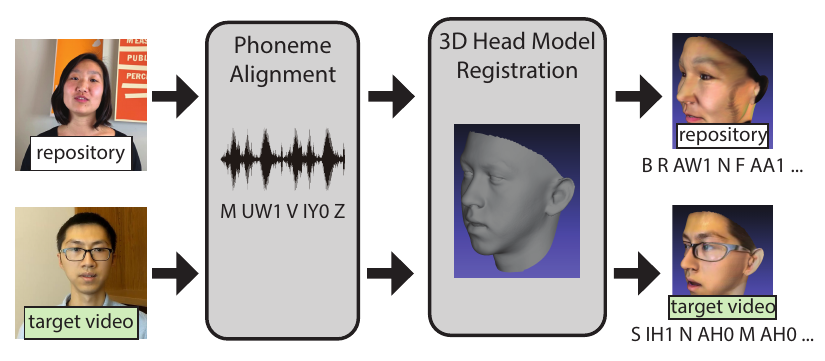}
  \vspace{-1.5em}
  \caption{Our preprocessing pipeline annotates both the source repository and the target video with phonemes and registers a $254$-parameter 3D head model to each frame of each video.
%    Overview for the preprocessing pipeline.  We preprocess both the
%    repository and the target video with phoneme alignment and 3D head
%    model registration.  This gives us the phoneme label, start and
%    end time for every phoneme in the videos, as well as $254$ face
%    parameters at each frame.
  }
    \vspace{-1em}
  \label{fig:preprocessing_overview}
\end{figure}

\begin{figure*}
  \vspace{-1em}
  \centering
  \includegraphics[width=\textwidth]{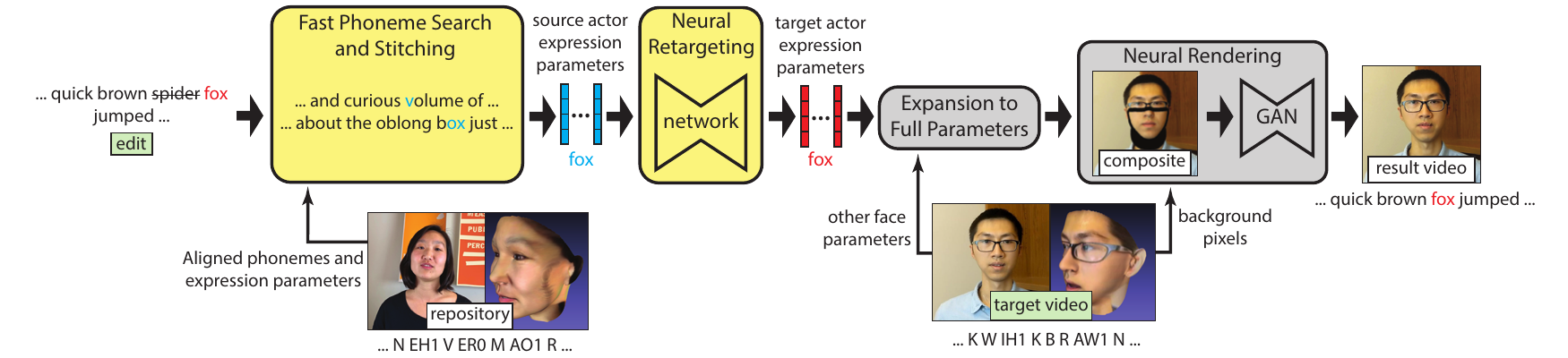}
  \vspace{-2em}
  \caption{Our synthesis pipeline adapts the pipeline of
    Fried~\etal~\shortcite{Fried_2019} by introducing a fast phoneme
    search and stitching step (yellow), and a self-supervised neural retargeting step (yellow). 
    % \maneesh{Only ``f'' should be highlighted in word ``fog''. Might be clearer  to use a ``v'' word rather than an ``f'' word to stress the fact that we are considering visemes, not just phonemes.}\david{I'm trying to show overlapped subsequences here: fog for F-AH-K and ox for AH-K-S. But maybe there's no need to delve that deep in details here?} \maneesh{Showing the overlap makes the figure way too confusing. We should go back to what we had earlier.}
    The input to our pipeline is a target video (green) and an edit (green) -- here
    changing the word ``spider'' to ``fox''.
    Our fast phoneme search
    finds phonemes in the repository that visually match the desired edit 
    -- here the ``v'' in volume and the ``ox''
    in box.  We then stitch together the corresponding facial
    expression parameters of the 3D head model for the source
    repository actor, and use a novel neural retargeting model to
    translate those parameters into those for the target actor.
    Next, we expand the retargeted expression parameters to the full
    face parameters (e.g. pose, illumination) for the target actor.
    Finally we render photorealistic frames from the parameters using
    a neural rendering approach that first composites the lower part
    of the face rendered from the 3D head model, with background
    pixels from the original target video and then uses a generative
    adversarial network (GAN) to map the composites to photorealistic
    frames.  }
  \label{fig:synthesis_overview}
\end{figure*}

\section{Method}
\label{sec:method}

Given a short talking-head video of a {\em target actor} (often 2-3
minutes in length), and an edit of the video transcript, our system
synthesizes new video of the target actor matching the edit. An
edit is specified as a replacement of one continuous sequence of words
in the original transcript with a new sequence of words.  Since a
short target actor video is unlikely to contain all the lip motions
necessary to convincingly synthesize the sequence of phonemes in the
edit, we leverage a large {\em repository} of video from a different,
{\em source actor}.
%that contains the most common phoneme combinations
%(coarticulations) in English.
%
%Specifically, we pre-capture an hour of
%a source actor speaking the TIMIT corpus\,\cite{garofolo1993timit} and
%retarget the lip motions of the source actor to the target actor
%during synthesis.
Specifically, we pre-capture an hour of a source actor speaking the
TIMIT corpus\,\cite{garofolo1993timit} which includes the most common
phoneme combinations (coarticulations) in English and we retarget
their lip motions to the target actor during synthesis.

Our approach for quickly synthesizing the edited result is based on
the approach of Fried et al.\,\shortcite{Fried_2019} but involves
several critical modifications.  As in Fried et al.,
%\ohad{wording a bit confusing, since Fried19's repository is not the
%same as ours (different person)},
our {\em preprocessing pipeline} (\Cref{fig:preprocessing_overview})
annotates both the repository and target videos with phonemes and
registers a parametric 3D head model to the face in each frame of each
video. Our {\em synthesis pipeline} (\Cref{fig:synthesis_overview})
provides a new, fast phoneme search algorithm that finds subsequences
of phonemes in the source video that match the desired edit. It then
stitches together the corresponding parameters of the 3D head model
for the source actor across subsequence boundaries to smooth the lip
motions.  We introduce a new self-supervised neural
retargeting step that adapts the parameters representing the lip
motion of the source actor to those of the target actor and blend the
resulting parameters into the target video.  Finally we render
photorealistic frames from the parameters using neural rendering\,\cite{tewari2020state}.
%\maneesh{cut next sentence for space?}\david{No objections.}
%The speed of our synthesis
%pipeline enables iterative editing and our tool provides additional
%controls for refining lip motions, inserting non-verbal mouth gestures
%and manipulating speaking style. 

We briefly summarize how we adapt each step in Fried et al.'s
pipelines to our problem in Sections~\ref{sec:preprocessing_overview}
and~\ref{sec:synthesis_overview}. We then present the details of our
new algorithms; fast phoneme search and stitching in \Cref{sec:search} and neural retargeting algorithm in
\Cref{sec:retargeting}.  In~\Cref{sec:interaction}, we
describe the iterative refinement and performance controls enabled by
our approach.

%% \maneesh{This high-level summary of the approach might be useful for
%%   the intro or abstract.}  We achieve this by representing the face
%% shape at each frame of the video with parameters for a 3D head model,
%% then synthesizing a time series of parameters that animates the face
%% to speak the new words, and finally rendering the parameters into
%% photorealistic frames of a talking head to replace the frames that
%% speak the original words.

\subsection{Preprocessing Pipeline}\label{sec:preprocessing_overview}
Our preprocessing pipeline annotates each frame of the repository and
the target videos in two main steps, (1) phoneme alignment and (2)
registration of a parametric 3D head model. These resulting phoneme
and face parameter annotations are used by our synthesis pipeline to
establish correspondences between the target video and source
repository.  Note that the repository is only annotated {\em once} and
the resulting {\em annotated repository} is then bundled as part of the
system. In contrast, the target video must be annotated each time a
new target video is given as input.

\paragraph{Phoneme Alignment}
The phoneme alignment step takes as input a video (repository or
target) paired with its text transcript, and computes the identity and
timing of the phonemes in the video.  Specifically, we use
P2FA\,\cite{yuan2008p2fa,rubin2013p2fa} to convert the transcript into
phonemes and align them to the audio speech track of the video.  This
produces an ordered sequence $V=(v_1, \ldots, v_n)$ of phonemes, where
each phoneme $v_i$ contains its name, start time and end time.
% We manually fix gross misalignments if they occur (about 10 instances in
%an hour-long video). \maneesh{Delete point about misalignments for space.}
If the transcript is not available, we can
obtain one using a transcription service such as Google Cloud
Speech-to-Text\,\shortcite{googlecloudstt}, or
rev.com\,\shortcite{rev}.

\paragraph{3D Head Model Registration}
We fit a parametric head
model\,\cite{blanz1999facemodel,ThiesZSTN2016a} to each frame of
video using a monocular head
tracker\,\cite{garrido2016reconstruction}.  At every frame, the fitted
model includes 80 parameters for 3D facial geometry, 80 for facial
reflectance, 3 for head pose, 27 for scene illumination and 64 for
face and lip expressions.  In the fitting procedure we hold the facial
geometry and reflectance parameters constant across all the frames of
the same actor, but we allow the pose, illumination and expression to
vary across time.

\subsection{Synthesis Pipeline}
\label{sec:synthesis_overview}

Our synthesis technique is based on matching phonemes in the edit
to phonemes in the repository. Therefore, we first convert the input
text of the edit from words into a sequence of phonemes $W=(w_1,
\ldots, w_m)$, where each phoneme contains its name, start time and
end time.
% \maneesh{I shorted prev. 2 sentences. Orig. longer version is in comments.}
%Our synthesis pipeline takes a target video and an edit as input, and
%generates a video of the target actor that matches the edit.  Since
%our synthesis technique is based on matching phonemes in the edit to
%phonemes in the repository, we start by converting the text of the
%edit into a sequence of phonemes $W=(w_1, \ldots, w_m)$, where each
%phoneme contains its name, start time and end time.
%
Specifically, we convert the edit into audio using either
%\maneesh{either a library of voice recordings of the target actor},
text-to-speech voice
synthesis\,\cite{wavenet2016,googlecloudtts} or voice cloning techniques\,\cite{kumar2019lyrebird,lyrebird}, 
and then apply P2FA\,\cite{yuan2008p2fa,rubin2013p2fa} to
time-align the resulting speech to the phonemes of the edit.
Note that our synthesis algorithm only uses the timing of the phonemes
and does not use any other aspect of the synthesized speech audio
signal.
%\maneesh{Do we need prev sentence?}
%
%% Specifically, we use either
%% text-to-speech\,\cite{wavenet2016,googlecloudtts} or voice
%% cloning\,\cite{jia2018sv2tts,jin2017voco,kumar2019lyrebird} to convert
%% the edit into audio and then apply
%% P2FA\,\cite{yuan2008p2fa,rubin2013p2fa} to time-align the resulting
%% speech to the phonemes of the edit.
%
% If the user later captures a recording of the target actor saying the
% new content, they can re-run our fast synthesis pipeline with the new
% timings obtained from the real recording. \maneesh{May want to update
%   last line depending on how we talk about real-voice results.}
% \david{If the user is the target actor, they can record speaking the
%   new content in real time and use our fast synthesis pipeline with
%   timings from the real recording.} \maneesh{Previous sentence is the
%   more likely scenario and should be combined with the sentence about
%   recording the actor later.}
% \maneesh{Can next sentence (not highlighted) replace sentence starting at ... If the user later captures ...}
If the user has access to the audio of the target actor saying the new
content (either from a prerecorded library of the actor's speech or
recorded by the actor in real-time during editing) they can run
our fast synthesis pipeline with the timings obtained from the
real-voice recordings.

\paragraph{Fast Phoneme Search and Stitching}
The fast phoneme search and stitching step is designed to quickly find the best
subsequences of phonemes in the repository and then stitch together the corresponding
expression parameters of the source actor 3D head model, in order to produce the
the edit $W$ we wish to synthesize.
% \maneesh{We need to say more about the stitching as we did in the previous version of the paper -- from what is written here it is unclear what stitching does. Need to say it ... stitches together the corresponding expression parameters of the source actor 3D head model to produce the edit $W$ we wish to synthesize....
  % while conforming to the phoneme timing specified in the edit $W$ we wish to synthesize.}
Our new algorithm 
operates three orders of magnitude faster than
Fried~\etal~\shortcite{Fried_2019} and finds the best repository
subsequences in seconds rather than hours. We present this fast
algorithm in detail in~\Cref{sec:search}.

\paragraph{Neural Retargeting}
The retargeting step converts a sequence of expression parameters for the source actor into those for
the target actor.
We introduce a
learned retargeting model, trained in a self-supervised manner from
corresponding pairs of 
repository and target video sequences and transforms the expression
parameters as detailed in~\Cref{sec:retargeting}.
%
%% Our retargeting method needs only a few minutes of video from target
%% actor and significantly lowers the data requirement on the target
%% video to about 2--3 minutes for high-quality results.  \ohad{we can
%%   add: "compared to, e.g., the method of Fried~et~al. which requires
%%   ~1.5 hours"}.
%
The result is a sequence of target actor
face expression parameters corresponding to the edit.

\paragraph{Expansion to Full Parameters}
Next, we combine the synthesized target actor expression parameters
with geometry, reflectance, pose and illumination parameters from the
input target video to 
produce a sequence of full face parameters for the target
actor corresponding to the edit.
%Next, we augment the newly synthesized target actor expression
%parameters with head pose and illumination parameters from the target
%video, to produce a sequence of full face parameters for the target
%actor.
Specifically, we take an interval of frames around the edit location
in the target video, retime it to account for the duration of the
edit, and use the geometry, reflectance, pose and illumination
parameters from the retimed interval.
%
%% \maneesh{I edited the
%%   next sentence, but I'm not sure I really understand what we are
%%   doing. I'm also not sure we need to include it as it is a detail.}
%% To ensure that the synthesized expression parameters fit seamlessly
%% back in the target video, we smooth the expression parameters at both
%% ends of the edit boundary with respect to the original expression
%% parameters in the target video.

\paragraph{Neural Rendering}
The neural rendering step takes the sequence of full face
parameters for the target actor and first generates a composite image
in which the lower face region is a rendering of the 3D head
model, while the upper part of the head and the surrounding background
are from the original target video, but retimed to match the length of
the edit. It then uses a GAN trained on the target video to
complete the image-to-image translation from composite image to
photorealistic frame.

%% The neural rendering step produces the final frames corresponding
%% to the edit by synthesizing each frame from the full face
%% parameters for the target actor.  We produce an intermediate
%% composite image by masking out the lower face region in the
%% original frame, replacing it with the synthetic render of the lower
%% face of the 3D head from the new face parameters, and finally use a
%% GAN (based on Fried~\etal~\shortcite{Fried_2019}) trained on the
%% target video to complete the image-to-image translation from
%% composite image to photorealistic frame.

\input{method-search}

\input{method-retargeting}

\input{method-interaction}
\input{method-runtime.tex}

%%  LocalWords:  photorealistic retargeting retimed coarticulations
%%  LocalWords:  viseme coarticulation retargeted datasets TIMIT PCA
%%  LocalWords:  dataset renderer Preprocessing misalignments Garrido
%%  LocalWords:  crowdsourcing reflectance blendshapes et retarget al
%%  LocalWords:  preprocessing visemes subsequence Ohad RNN relu pre
%%  LocalWords:  regularizer Hyperparameters minibatch Tensorflow len
%%  LocalWords:  memoization bigram unigram bigrams runtime retime
%%  LocalWords:  retiming subsequences lbl GAN iteratively
%%  LocalWords:  levenshtein googlecloudstt

%% file: method-search.tex
\subsection{Fast Phoneme Search and Stitching}\label{sec:search}
Our synthesis pipeline takes an edit $W$ specified as a sequence of
phonemes with timings $(w_1, \ldots, w_m)$ and starts by finding
matching subsequences of video in the source repository $V$, that can be
combined to produce $W$.
% \david{Do we talk about vanilla Fried'19 here or
% go straight to context-aware search? I think it's more clear to recap Fried19 as is done here,
% and talk about the improvement later, although that might make the novelty appear more incremental.}
% \maneesh{I think we just want to lightly lace it into what is already here. Isn't it just that we partition the edit into slightly overlapping subsequences?}
More precisely, we partition the edit $W$
into phoneme subsequences $(W_1, W_2, \ldots, W_k)$ and for each
subsequence $W_i$ find its best match $V_i$ in the repository $V$.
Fried~\etal\,\shortcite{Fried_2019} use a brute-force method that
considers all possible partitions $\mbox{split}(W)$ of the edit $W$,
and all possible matches with subsequences of $V$ to find $(V_1, V_2,
\ldots, V_k)$ that minimizes the objective:
$$C(W,V)=\min_{\substack{
    (W_1,W_2,\ldots,W_k)\in\mbox{split}(W)\\ (V_1,V_2,\ldots,V_k)
}}\sum_{i=1}^k \cmatch(W_i,V_i) + C_{\mbox{len}}(W_i)$$ where
$\cmatch$ is a custom Levenshtein edit
distance\,\shortcite{levenshtein1966} between two phoneme subsequences
that takes into account the phoneme label, the viseme label and the
timing difference, and $C_{\mbox{len}}$ penalizes short subsequences.
In order to find subsequences that transition well at their
  endpoints, during the search, we expand each subsequence $W_i$ by a
  single phoneme on either end. Thus, adjacent subsequences overlap by
  two {\em context} phonemes.  We find that this new {\em context expansion}
  approach better captures co-articulation effects between the
  subsequences, than the algorithm of Fried et al. which does not use
  context expansion (see user studies in \Cref{sec:user_study}).
%  Note that unlike Fried et al, we
%  expand each $W_i$ by one phoneme on either side during this search so that adjacent subsequences overlap by a phoneme.
%  In practice we have found that we
%  obtain better transitions between the matching phoneme subsequences using
%  this {\em context aware approach} as the overlapping phonemes better
%  account for coarticulation effects.}
%\david{In order to find subsequences that transition better from one into the next, we
%employ a new \emph{context aware} scheme where each $W_i$ is expanded to include the phoneme before and after.}
%\maneesh{Prev sentence is a little unclear -- I put an alternate version earlier. What do you think?}

We further modify Fried et al.'s search algorithm in three key
ways to obtain a speedup of over 3 orders of magnitude: (1) we propose
a fast alternative to to the Levenshtein distance, (2) we reduce the
size of the search space on $W_i$ and, (3) given $W_i$, we use a
viseme-based indexing scheme to quickly find the optimal $V_i$.
%\maneesh{Move context awareness to here as ... In addition, unlike Fried et al., we expand each subsequence $W_i$ by one phoneme (i.e. a context phoneme) on either side so that the resulting matches from the repository overlap by a phoneme on each end. In practice we have found that we obtain better transitions between the phoneme subsequences using this {\em context aware approach} as the context phonemes better handle coarticulation effects.}
Finally, we stitch together source actor expression parameters corresponding to the $V_i$s
to produce a single coherent sequence of expression parameters.
% \maneesh{Probably need to mention  stitching in a short sentence here as we will not how a paragraph or two on that.}

\paragraph{(1) Fast alternative to edit distance}
The full Levenshtein edit distance allows substitution, insertion and
deletion of phonemes when computing $\cmatch$.  However, we have observed that when the
matching subsequences between the edit $W$ and the repository
video $V$ contain phoneme insertions or deletions, the final
synthesized video appears out-of-sync with the audio; it either
contains extraneous mouth motions due to phoneme insertion, or it 
misses mouth motions due to deletion. In practice we find it is beneficial
to disallow insertions and deletions and only allow phoneme
substitutions.  Given a subsequence $W_i$ of the edit $W$, this approach
forces $\cmatch$ to only consider subsequences $V_j$ of the
repository $V$ that contain the same number of phonemes as $W_i$.  We
can therefore replace the Levenshtein distance with the sum of
element-wise substitution cost which requires linear time in the
number of phonemes rather than the quadratic time required for
computing the full Levenshtein distance\,\cite{wagner1974string}.

\paragraph{(2) Reduce search space for partitioning}
The brute force search considers all possible partitions of $W$ into
$(W_1,\ldots,W_k)$. But, an extremely long edit subsequence $W_i$ is
unlikely to have a good match with a repository subsequence $V_i$.
Thus, we can reduce the search space of possible partitions by capping
the maximum length of the $W_i$'s to $L$.  In our experience, 
99\% of the matches found by the brute force search
are of length $6$ or less, and we therefore set $L=6$.
This approach reduces the number of partitions to search.
More importantly, it typically reduces the number of distinct $W_i$'s we need to
consider by over an order of magnitude, especially when $W$ the full edit
sequence is itself very long.

\paragraph{(3) Viseme-based index to search repository}
For each edit subsequence $W_i$ we consider in our search space, we
must find the optimal $V_i$ in the repository with respect to
$\cmatch$.  Instead of checking all possible subsequences in the
repository, we impose an additional constraint on $V_i$ that restricts
the set of $V_i$ we consider to only the most likely match candidates
and allows us to build an index structure on the set of $V_i$ to
retrieve the likely candidates quickly.

As in Fried et al., our $\cmatch$ cost function considers phonemes to
match when they appear visually similar -- that is, their
corresponding visemes match. By imposing the restriction that $V_i$
start with the same viseme n-gram as $W_i$ we can pre-compute an index
for the repository using viseme n-grams as the key and the location of
the n-gram in the source repository video as the value.
At search time, we look up all possible candidate $V_i$'s using this
index and only compute the $\cmatch$ for them.  While this indexing
approach speeds up the search significantly it also reduces the space
of subsequence matches the search considers. In general, the longer the n-gram
key, the stronger the reduction and the more likely it is that no good
match will be found. In practice, we find that using a bi-gram index
best balances this trade-off between search speed and result quality.

\paragraph{Stitching}
% \maneesh{updated section.}
After we find the best matching phoneme subsequences $V_1,\ldots,V_k$
from the repository, we look up the expression parameters of the
source actor's 3D head model corresponding to each phoneme, and
linearly re-time them to match the phoneme durations specified in the
edit.  We then stitch together adjacent subsequences by first linearly
blending the the expression parameters across the overlapping context
phonemes and then applying a Gaussian filter over a window of 4 frames
around the transition boundaries between the subsequences to further
smooth the transition.

We have found in practice, that errors in tracking expression
parameters, timing misalignments and our linear blending can sometimes
fail to fully close the mouth of the 3D head model at the beginning of
\textbackslash m, \textbackslash b, and \textbackslash p phonemes. However,
proper mouth closures for these phonemes is crucial for producing
perceptually realistic results\,\cite{agarwal2020}.  We therefore force the desired
mouth closure by linearly blending in a closed-mouth expression from
the repository at the beginning of all \textbackslash m,
\textbackslash b, and \textbackslash p phonemes with a default length of 2
frames.  Note that the closed-mouth expressions are manually annotated
once during repository preprocessing and we automatically use the one
closest to the parameter values at the point of insertion.

%% \david{Phoneme misalignments and the blending procedure can sometimes
%%   fail to fully close the mouth of the 3D head model, such as at the
%%   beginning of \textbackslash m, \textbackslash b, \textbackslash p
%%   phonemes.  Proper mouth closures at the right time are crucial to
%%   the perceived realism of results \cite{agarwal2020}.
%% We therefore
%%   force the desired mouth closure by linearly blending in a
%%   closed-mouth expression from the repository at the beginning of all
%%   \textbackslash m, \textbackslash b, \textbackslash p phonemes with a
%%   default effect radius of 2 frames.  Note that the closed-mouth
%%   expressions are manually annotated once during repository
%%   preprocessing and we automatically use the one closest to the
%%   parameter values at the point of insertion.  }
% \david{Mouth closures at \textbackslash m, \textbackslash b, \textbackslash p phonemes are crucial to the perceived realism of results \cite{agarwal2020},
% but misalignments of phonemes and the blending procedure can sometimes fail to close the mouth of the 3D head model fully.
% We therefore force the desired mouth closure by linearly blending in a manually-annotated closed-mouth expression from the repository at the beginning of all \textbackslash m, \textbackslash b, \textbackslash p phonemes. 
% }
% \maneesh{Needs more detail and also a better lead in -- talk about the misalignment problem, then about the realism. How did we obtain the closemouthed frames. How long do we blend them in for? }

\vspace{0.5em}
Overall, the fast search strategy reduce the runtime of
the phoneme search process by three orders of magnitude compared to
the brute-force approach.  It takes around $5$ seconds to find
and stitch snippets $V_1,\ldots,V_k$ for an edit $W$ of $20$
phonemes in an hour-long repository.

%% file: method-retargeting.tex
\subsection{Neural Retargeting}\label{sec:retargeting}
Given a stitched-together sequence of face expression parameters for
the source actor in the repository, the goal of retargeting is to
generate a matching sequence of expression parameters for the target actor.
%\ohad{should we add a sentence explaining why these two sets of
%  parameters are different in the first place?}
We have developed a self-supervised neural network model for retargeting
and in Section~\ref{sec:eval_algo_data} we show that
using a neural network for retargeting produces
%%with self-supervision based on
%%corresponding pairs repository and target video 
%%visemes
%%produces
higher-quality results than baseline
methods such as directly copying source actor expression parameters to
the target actor, or applying a linear retargeting model.

%as we will show in \Cref{sec:eval_retarget}.  \maneesh{May want to
%  move the comparison to end of this section.}

\begin{figure}
\centering
\includegraphics[width=\linewidth]{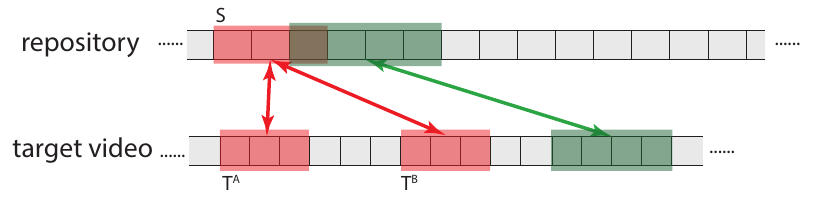}
\caption{To automatically build training data for our neural
  retargeting model, we consider a sequence of phonemes $S$ (red) in
  the repository and find up to two best matches $T^A$ and $T^B$ (red)
  in the target video.
%
%
%  Phonemes in the video are represented as cells. For a
%  sequence $S$ (red) in the repository, we find up to two best matches
  %  $T^A$ and $T^B$ (red) in the target video.
  We then find
  matches for the next repository sequence (green) starting at
  the last phoneme of the previous sequence.  The overlap allows
  us to capture the phoneme transition out of the final phoneme in the red subsequence.}
\label{fig:build_correspondence}
\end{figure}

\paragraph{Self-supervised training data}
%\david{recent rewrite}
To train our retargeting network we require sequences of expression
parameters for the source and target actor that correspond to one
another with respect to their mouth motions.  Assuming that uttering
the same sequence of visemes will produce similar mouth motions, we
automatically construct corresponding pairs of training data by
finding the longest matching sequences of phonemes between the source
repository video and the target video, as follows.

Since we apply our retargeting model to a stitched-together
subsequences of phonemes that can come from anywhere in the source
repository, we would like the training sequences to cover as much of
the repository as possible.
Therefore, we start from the first phoneme $s_1$ in the
repository, and find the longest sequence $S=(s_1,\ldots,s_k)$ for
which there is at least one corresponding sequence
$T=(t_1,\ldots,t_k)$ in the target video where $t_i$ and $s_i$ belongs
to the same viseme group (i.e. phonemes that require the same lip expressions
are in the same viseme group).
We take up to two best matches in the
target video with respect to $\cmatch$ score defined in
\Cref{sec:search}, and call them $T^A=(t^A_1,\ldots,t^A_k)$ and
$T^B=(t^B_1,\ldots,t^B_k)$.  We add the pairs $(S,T^A)$ and $(S,T^B)$
to the set of phoneme sequences in correspondence, and continue the
scan through the source repository at $v_k$, until we finish scanning
through the entire repository for subsequence matches
(\Cref{fig:build_correspondence}).  To quickly find the best $T^A$ and
$T^B$ sequences in the target video, we apply fast search techniques discussed
in \Cref{sec:search}.
Finally, to convert each resulting phoneme sequence pair into a parameter
sequence pair, we linearly retime the target video phoneme interval to
the corresponding repository interval and similarly interpolate the
% corresponding
expression parameters.
The retargeting model is trained once for each new target video.
%\maneesh{Are we now doing this in a context aware way? Also have we
%  tried using more than the top two matches? I wonder if considering
%  more matches would help us get the mouth to open when it should.}
%\david{
%  Our neural retargeting step is unchanged from January submission.
%}
%\maneesh{ok, but in the paper here we say we use the score defined in section 3.3 and that score is described in a context-aware way so the assumption  would be that we would be doing this here in a context aware way. My sense is that we should try the context aware approach if we are not doing it that way now.}

%% Note that because of coarticulation effects, the phonemes at the start
%% and end of a sequence may not produce similar mouth motions in the
%% corresponding repository and target videos. We therefor discard first
%% half of the frames in the first phoneme interval and last half in the
%% last interval, effectively discarding all single-phoneme sequence
%% pairs. \maneesh{This is a detail we might consider removing later.}

\begin{figure}
\centering
\includegraphics[width=\linewidth]{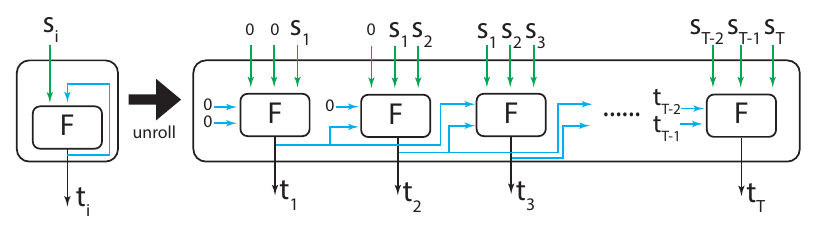}
\caption{The retargeting network with the unrolled recurrent unit $F$, which looks two frames back at each time step and uses the previous two outputs as its state.
The diagram shows how the RNN processes the input of $T$ frames of source actor parameters $(s_1, s_2,\ldots,s_T)$ and produces the prediction for $T$ frames of target actor parameters $(t_1, t_2,\ldots,t_T)$.}
\label{fig:retargeting_network}
\end{figure}

\paragraph{Neural Network Architecture}
%\david{recent rewrite}
We employ a recurrent neural network (RNN) manually unrolled for $T$
time-steps to encode the temporal dynamics of the facial expressions
and regress from source parameters to target parameters
(\Cref{fig:retargeting_network}).  The resulting network takes as
input $T$ frames of $64$ expression parameters for the source actor,
and outputs $T$ frames of $64$ parameters for the target actor.  At
the core of the network is the recurrent unit that is made up of $3$
fully connected layers of $1024$ nodes with relu activation. The
inputs to the recurrent unit are $64$ expression parameters of the
current time-step, as well as the parameters and outputs of the
recurrent unit for the previous $H$
%\maneesh{I am a little lost. Is H shown in the figure at all? Feels
%like it should be.} \david{The figure only shows the case for $H=2$
%which is what we used.}
time-steps.  To facilitate
learning of deviations from the identity transformation, the output
layer of the recurrent unit with $64$ nodes and no activation produces
the residual values that are added to the input source parameters
element-wise to obtain the prediction for the target parameters for
the current time-step. We zero-pad the unavailable inputs at the
first $H$ time-steps. We empirically found that setting $H=2$ and
$T=7$ produced high-quality retargeting results.

\paragraph{Loss function}
%\david{recent rewrite}
Our loss function is a linear combination of a data term and a
temporal regularizer with regularization weight $\lambda$:
$$\mathcal{L} = \frac{1}{T}\sum_{i=1}^T \|F(s_i)-t_i\|_1 +
\lambda\|F^{(2)}(s_i)\|_2$$
where $s_i,t_i$ are $64$-dimensional vectors representing the $i$th time-step of the source and target parameters
respectively,
$F(s_i)$ is a $64$-dimensional vector of the predicted target actor parameters and
$F^{(2)}(s_i)$ is a $64$-dimensional vector that is the second temporal difference
vector, or acceleration, of the predicted values.
Empirically
we found that using an $L_1$ norm for the data term significantly outperforms
$L_2$ by generating more expressive motions and better preserving mouth closures.
The temporal regularization term is needed to make the network
predict temporally stable parameters.

\paragraph{Hyperparameters and training}
Since the network takes a fixed-size input of $T$ frames, we run a
sliding window on each matching parameter sequences to obtain the
training examples.  Experimentally we set the temporal window $T=7$
frames.  For training we set $\lambda=10$, and dropout rate at 25\%,
50\% and 25\% for the three layers in the recurrent unit,
respectively.  To train the network we use stochastic gradient descent
with the Adam solver\,\cite{Kingma2014Adam} and set an initial learning
rate of $0.0002$ with an exponential decay rate of $0.5$.  We employ
gradient clipping\,\cite{pascanu2013clipping} to avoid exploding
gradients.  We train the network with minibatch size $100$ and
training typically converges within $100$ epochs.

\paragraph{Inference}
At inference time, we convert a sequence of source actor expression
parameters into target actor parameters.  Since our retargeting model
accepts fixed-size $T$ frames of input and produces $T$ frames of
output, we run a sliding window of length $T$ over the new sequence of
source actor expression parameters at inference time. Each frame is
covered by exactly $T$ such sliding windows.  In order to obtain a
more temporally stable output, at each frame we average the $T$
outputs produced by those $T$ sliding windows as the final output of
the frame.  The result is a synthesized sequence of target actor
expression parameters that animate the face to speak the new content
of the edit with the desired timing.
We then proceed to expand these expression parameters into parameters
for the whole head and use neural rendering to generate the video frames
(\Cref{sec:synthesis_overview}).

\vspace{0.5em}
Training our retargeting model typically requires 2--3 minutes of
target actor video speaking arbitrary speech to produce high-quality
synthesis results. Retargeting allows our tool to leverage the large
repository of controlled source actor video (speech consists of TIMIT
sentences) to generate the target actor lip motions and opens our tool
to many practical applications where large amounts of controlled
target actor video are not available.

% LocalWords:  retargeting na ve eval retarget uncanniness viseme RNN
% LocalWords:  visemes subsequence bigram retime coarticulation Ohad
% LocalWords:  relu regularizer th Hyperparameters minibatch TIMIT NN
% LocalWords:  Tensorflow cmatch subsequences

%% file: method-interaction.tex
\subsection{User Controls} \label{sec:interaction}
%As we will show in Section~\ref{sec:runtime} our synthesis pipeline
%can generate video corresponding to a typical edit in about
%40 seconds.
This speed of our synthesis pipeline opens
the door to interactive user controls for iteratively refining the
edit and further manipulating the the facial performance.

%% \begin{figure}
%% \centering
%% \includegraphics[width=\linewidth]{figs/ui.pdf}
%% \caption{The user interface of our tool.}
%% \label{fig:ui}
%% \end{figure}

\paragraph{Refinement Control: Smoothing jumpy transitions}
Our synthesis pipeline stitches together different subsequences of
expression parameters from the source repository by smoothing over a
window of 4 frames around the transition boundary
(Section~\ref{sec:search} Fast Phoneme Search and Stitching).  At
times however, some transitions may still appears jumpy even after
this smoothing.
%\maneesh{the smoothing introduced by} retargeting model inference.
%\maneesh{Unclear why we are mentioning retargeting inference when we
%  first talked about stitching.  I added something about smoothing
%  introduced, but I don't really understand why retargeting would
%  introduce smoothing.}
We allow the user to inspect the result and
further refine it by manually specifying the interpolation radius (in
number of frames) at user-specified transition boundaries to better smooth out visibly
jumpy transitions.

\paragraph{Refinement Control: Adjusting mouth closure}
%\david{Need to rewrite this paragraph. Instead of mentioning forced
%  MBP closure in previous sections, an alternative is to introduce it
%  here. Since we force-blend MBP shapes automatically already, this
%  control is really for adjusting the strength of the blend.}
%\maneesh{I don't think we need to be so detailed about this and can keep it mostly as is. We should discuss.}
%\david{Most content here are now in \Cref{sec:search} when introducing forced MBP mouth closures.}
% \maneesh{Updated section... May want to call it ``Adjusting mouth closure'' I'm not sure. Old version of paragraph is in comments.}
As noted in Section~\ref{sec:search} mouth closure on \textbackslash
m, \textbackslash b and \textbackslash p phonemes is crucial for the
mouth motions to appear realistic. Thus our stitching procedure
automatically inserts 2 closed mouth frames at the beginnings of these
three phonemes to ensure the mouth closes correctly for them. We
further allow users refine any synthesized result by
extending (or reducing) the length of the inserted closed mouth
frames.
%
%% Our automated pipeline sometimes fails to fully close the mouth of the
%% target actor properly, such as at the beginning of \textbackslash m,
%% \textbackslash b, \textbackslash p phonemes, due to either inaccurate
%% phoneme alignment in the repository or smoothing from subsequence
%% stitching.  We allow the user to inspect the result and further refine
%% it by blending in a manually-annotated closed-mouth expression from
%% the repository at the specified location with different effect radius
%% before the retargeting.  Note that we manually annotate the closed
%% mouth expressions in the repository once during repository
%% preprocessing, and later automatically use the one closest to the
%% current parameter values at the specified location when the user
%% inserts it.

\paragraph{Performance Control: Inserting mouth gestures}
%\david{recent rewrite}
Users can also insert non-verbal mouth gestures (e.g. a smile)
into an edit. To enable such performance control we manually annotate
mouth gestures including {\em rest, closed-mouth smile, regular teeth-showing
smile, big open-mouth smile, sad, scream, mouth gesturing left} and
{\em mouth gesturing right,} in the repository video.  These segments can
then be retrieved by our fast phoneme search just like any other
phoneme annotation.
%\maneesh{I think showing some more of these in our results would help
%  -- the mouth left and right gestures are a bit uncommon -- it would
%  be great to think of examples that make sure of scream, and sad.}
%\maneesh{Make sure all expressions are shown in some example in the supplemental materials and show generalization -- that is input target video did not contain a gesture but we can still synthesize it.}

Since the annotations are on the repository, this manual annotation only
needs to be done once during repository preprocessing.  Note however, that 
users do not label the target video with these mouth gestures and our
retargeting network is never explicitly trained with corresponding
pairs of mouth gesture frames between the repository and target
videos. Nevertheless, we have found that our retargeting network is
able to generalize to unseen expressions and produce good quality
expression parameters for the target actor.
%\maneesh{Would be good to show more results of just the gestures generalizing to new target actors.}

With these annotations, the user can add special {\em mouth gesture
  directives} like [smile] anywhere in their edit of the
transcript, and our tool constructs a ``generalized phoneme'' edit
sequence $W$ that contains phonemes and such directives.  Any mouth
gesture that appears in $W$ is given a default duration of $0.5$
seconds that the user can override with an explicit duration
e.g. [smile:1.5s].  We employ a special substitution cost in
$\cmatch$ described in \Cref{sec:search} for ``gesture phonemes'' that
is set to infinity for a non-match to ensure that we retrieve the
correct ``phoneme'' match for the gesture.  When there are multiple
candidates, $\cmatch$ takes duration into account and picks the
gesture with duration closest to the query.  The rest of the editing
pipeline (\Cref{sec:synthesis_overview}) is otherwise unmodified.
% \ohad{there are way too many ``quotation marks'' in this paragraph}

\paragraph{Performance Control: Adjusting speaking style}
%\david{recent rewrite}
Our tool allows the user to select a different speaking style for the
synthesized result by using a version of the repository with the
desired style. In addition to the default repository which captures a
``neutral'' speaking style, we have recorded an ``energetic''
repository of our source actor with more pronounced mouth movements,
and a ``mumble'' repository with significantly less mouth movements. Figure~\ref{fig:sessions} (third row) shows
frames from these alternative repositories.

Importantly, we do not have to retrain our neural retargeting model
(\Cref{sec:retargeting}) for each additional style repository.  We
train this retargeting model once using only the default neutral
repository.
We have found that our default retargeting model can extrapolate to
retarget subsequences of source actor face parameters retrieved from
other speaking style repositories of the same actor.  Moreover, the
other repository videos can be captured at different times, with
different background and the source actor can even be wearing
different clothes or have a different hairstyle. 
%\maneesh{probably should say this point earlier or reinforce in limitations section..}
Thus, our tool generates videos with different speaking styles by
using one of the alternative style repositories in the fast phoneme
search step, but leaves the remainder of the synthesis pipeline
unchanged.

%% file: method-runtime.tex
\subsection{Implementation Details}
\label{sec:runtime}
%\maneesh{I think this stuff on running time should either go in an implementation section before results -- and also show the source actor video, or somewhere in the results section.}
%\maneesh{I think we should call this implementation and open by describing the machine that we've implemented on -- NVIDIA GTX... Then give running times.}
We implemented the fast phoneme search in Python and both our neural
retargeting model and the GAN renderer in
TensorFlow\,\cite{tensorflow2015}.  The monocular head tracker and
renderer are written in C++ with shader language extensions.

In preprocessing the repository and target videos, phoneme
alignment takes one third of the video time, and face registration takes
110 ms per frame.  It takes 30 minutes to generate training data for
our neural retargeting model and another 30 minutes to train it on one NVIDIA
GTX 1080Ti.  Training the GAN for neural rendering takes 17 hours on
one NVIDIA Tesla V100.
%Preprocessing, retargeting model training and
%neural renderer training is done once for each target video.

In our synthesis pipeline, 
our fast
phoneme search requires 5 seconds for a 
typical edit of
5 words containing 20 phonemes.  Retargeting inference speed is 10K
fps.  Composite images are rendered at 12 fps and final GAN rendering
takes 7 fps on two NVIDIA GTX 1080Ti. All together, a typical 5 word edit takes
around $30$ seconds for the full video generation (\Cref{tab:runtime}).
%We have also
%profiled our tool on edits of varying length and the runtime summary
%is in \Cref{tab:runtime}.

\begin{table}\small
\centering
\begin{tabular}{cccSSS}
\toprule
{\#words} & {\#phonemes} & {\#frames} & {search (s)} & {render (s)} & {total (s)} \\
\midrule
1 & 4  & 24  & 1.51  & 9.96  & 12.39 \\
% 2 & 10 & 32  & 2.12  & 10.00 & 15.45 \\
3 & 15 & 49  & 2.67  & 12.94 & 20.30 \\
% 4 & 18 & 53  & 4.18  & 13.56 & 22.26 \\
% 5 & 23 & 66  & 6.21  & 16.74 & 28.25 \\
6 & 25 & 72  & 5.32  & 17.74 & 28.95 \\
% 7 & 27 & 71  & 6.68  & 17.70 & 29.89 \\
8 & 39 & 105 & 7.57  & 24.33 & 37.97 \\
% 9 & 45 & 109 & 9.58  & 25.58 & 41.43 \\
10& 49 & 134 & 11.60 & 31.19 & 50.60 \\
% 19& 71 & 174 & 15.41 & 39.21 & 63.02 \\
\bottomrule
\end{tabular}
\caption{Runtime of our tool on a variety of edit lengths.
Search time scales roughly linearly with the number of phonemes, and render time scales linearly with the number of frames.
Even for long edits of 10 words, our system can generate video in approximately a minute.}
\label{tab:runtime}
\vspace{-3em}
\end{table}

It should be possible to further reduce the feedback time by
parallelizing our synthesis pipeline.  Phoneme search could be
distributed where each worker job is responsible for searching a
fraction of the repository.  Both parts of the neural rendering step
-- forming the composite images from target actor head parameters and
applying the GAN to generate photorealistic frames -- are
parallelizable by distributing the frames.  \new{We have performed
  initial experiments on parallelizing the GAN rendering which is the main bottleneck in our pipeline. Distributing the GAN rendering
  across a cluster of 8 NVIDIA Tesla V100s
  achieved a rendering rate of 24fps, a 3.4x speedup from the original 7fps for this step. Note that this speed up rate includes image compression overhead. Overall this experiment cuts the end-to-end synthesis time from $40$
  to $20$ seconds for a typical 8-word sentence.}  Similarly parallelizing the other parts of the pipeline and using sufficient hardware we believe that the end-to-end video generation feedback time could
be reduced significantly.  Streaming the frames as they are ready
could also further reduce the latency from issuing an edit to seeing
the first frames of the result, enabling real-time interactive editing
sessions.
%\maneesh{Perhaps should say this more strongly and emphasize it more. Make the case that we should be able to get to interactive rates.}

% LocalWords:  retargeting GAN renderer shader preprocessing GTX
% LocalWords:  cccSSS parallelizing photorealistic parallelizable

%% file: results.tex
\section{Results}\label{sec:results}
%\maneesh{Need to decide if we will be referring to ``our tool'' or ``our system''. I think what we really have is a tool. Let's check with Kayvon on this.}
%\maneesh{Need to decide if we will primarily be using the term ``interactive'' or ``iterative''.}

\begin{figure*}
  \vspace{-1em}
  \centering  
\includegraphics[width=\textwidth]{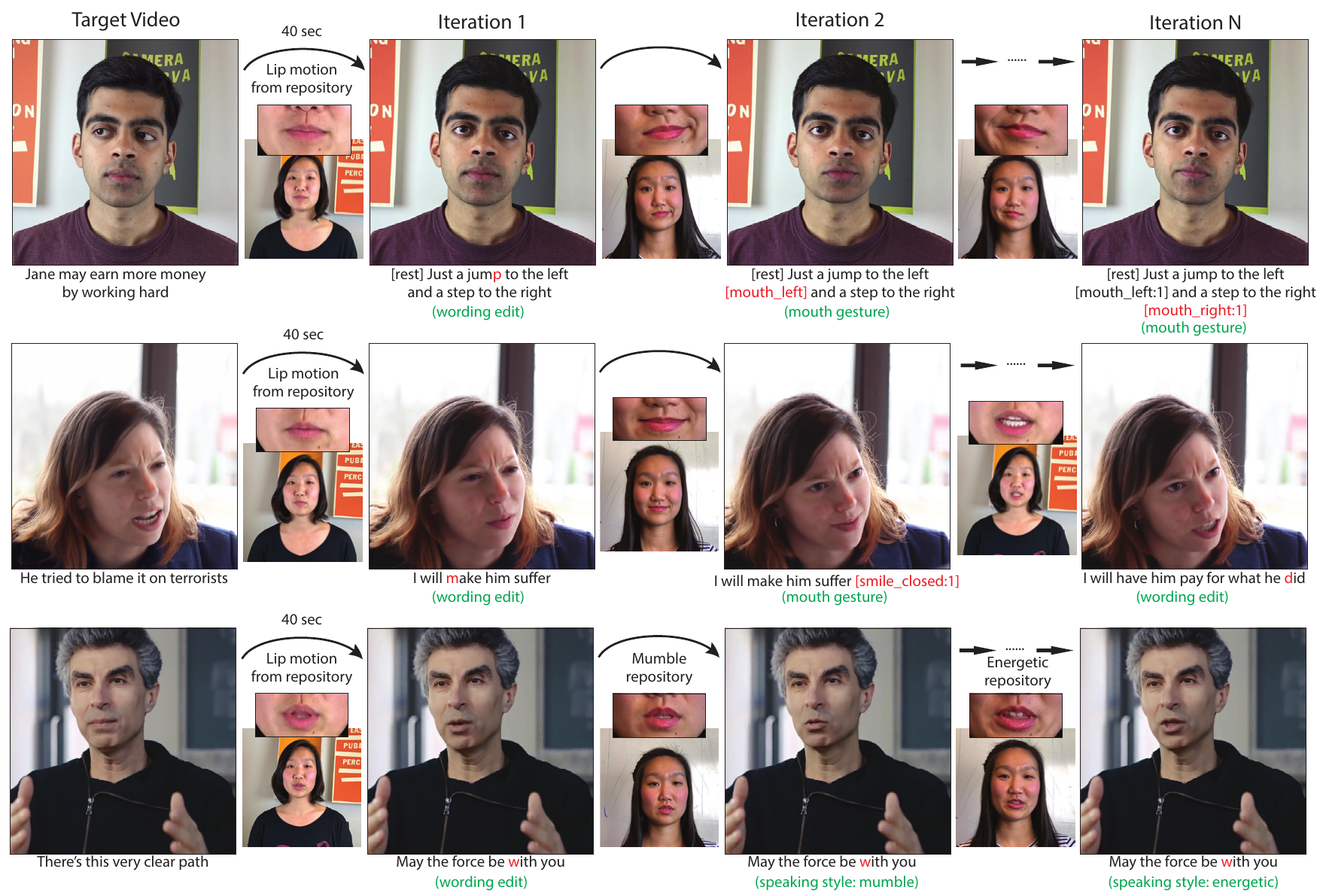}
\vspace{-2em}
\caption{Editing sessions facilitated by our tool on target videos of 1 to 5 minutes in length. In these sessions our tool lets
  the editor change wording, insert mouth gestures, refine mouth motions and manipulate speaking style. 
%  We present edit sessions on a variety of target videos of 1
%  to 5 minutes.  During these sessions, our tool enables the user to
%  perform various types of edits including wording edits, inserting
%  mouth gestures, refining mouth motions and changing speaking style.
  Here we highlight a few of these edits in each session.  Our tool
  finds the source actor mouth motions that conform to the specified edit
  operation and then retargets them to the target actor.
  %\Cref{sec:results} details the edit operations in
  %each session, and the
  Please see our supplemental materials and video for full
  video results as well as screen recordings of the edit sessions in our interface.  }
\label{fig:sessions}
  \vspace{-1em}
\end{figure*}

% \david{After the results are selected for the video, pick a subset to present in text here.}
\Cref{fig:teaser,fig:sessions} show examples of iterative text-based
editing sessions for a variety of target videos including recordings
of graduate students, YouTube video and a take from filming a dialog
scene. We encourage readers to watch the videos in our supplementary
materials to see how our text-based interface facilitate the
iterative editing workflow used in each of these sessions.

%\maneesh{We might want to describe these sessions as if a third person edited them --  could call the user the editor.} 
%\ohad{done. Please go over it and see if it sounds ok. One issue is that David is our only editor, so I kept saying "he" all over the place. I changed it to "they" in some sessions which I believe is still correct.}
%\maneesh{Also the session descriptions are a bit long. Can we tighten? ONe option is to say less about the iterations. let's discuss.}

%\maneesh{Perhaps we can tighten up these results descriptions as they are basically walking through the figures. But I'm not totally sure how best to tighten them.}

\paragraph{Session 1: Talking-head with glasses.}
%\maneesh{If we can we should use a heading that describes more about the type of video.}
Our first session works on a 2.5 minute target video
(\Cref{fig:teaser}).  The editor explores ways for the actor to parody Martin Scorsese and
express that Marvel movies are not to be considered real cinema.  They
first synthesize ``Marvel movies are not cinema'' from a resting pose.
Feeling it is too
blunt,
%and may cause a controversy, 
they slightly change
the wording to ``Marvel movies are not really cinema'', and eventually
settle on the firmer statement ``Marvel movies really are not cinema''. They then insert a smile at the end to soften the overall tone.
Our tool is able to produce results with synchronized mouth motions at each step.

\paragraph{Session 2: Talking-head with stubble.}
Our second session works on a 3.5 minute target video (\Cref{fig:sessions} first row).
The editor explores ways for the actor to give the instruction to start the time warp by jumping to the left then stepping to the right.
They first synthesize the instruction phrase, then add
gestures ``[mouth\_left]'' and ``[mouth\_right]'' to the corresponding
location in the dialog.  Next, feeling the gestures go too quickly, they lengthen the gestures to one second
each.
%, and finally smooth out a jumpy transition right after the
% gesture to the left. \ohad{maybe remove the jumpy transition removal from this one. We already described one, so I don't think it adds anything and it's not in the figure anyhow.}  
Our tool produces realistic video with mouth
motions synchronized to the audio and the gesture directives.

\paragraph{Session 3: Movie scene.}
Our third session works on a target video of a single take from a
dialogue scene (\Cref{fig:sessions} second row).  The target video is a
challenging one because it is only 1 minute long, and the actress
speaks for only 30 seconds in the take.  Our
tool nevertheless is able to synthesize compelling results for this
session.  In the session, the editor prototypes ways for the actress to express her hatred
towards the murderer of her sister's dog.  They first try ``I will make him suffer''.  Then for added creepiness, add a tight-lipped smile of
1 second duration at the end.  Finally, they settle on a less hostile line instead.
While the neural renderer struggles with the lack of
data to produce images as sharp as those from the previous two
sessions, our tool is still able to produce realistic and
synchronized mouth motions that give the user a good sense of how the
scene would look with the alternative line and gesture.

\paragraph{Session 4: Interview.}
Our fourth session works on a 5 minute excerpt of a YouTube video
(\Cref{fig:sessions} third row).  The editor explores different
delivery styles for the phrase ``may the force be with you''.
They first synthesize the phrase with the default repository, then
switch to a mumble style and finally to an energetic style.  While the
mouth movements match the audio, there is visibly less motion
with the mumble style and more motion with the energetic style.

%% \vspace{0.5em}
%% For complete video results, more editing session examples and more
%% synthesis results using both synthetic and
%% post-captured real-voice audio, please see the supplemental
%% material. \maneesh{I tried to get the idea of different ways of doing
%%   the audio here. But I'm not sure ``post-captured'' is the best way
%%   to say this but we need to make sure it is clear that the real voice was captured only at the end.}

% LocalWords:  retargets renderer

%% file: evaluation.tex
\section{Evaluation}

% \maneesh{Updated paragraph.} 

%% To evaluate our tool we analyze the quality of the synthesized video
%% as we vary the phoneme search and stitching method, and the
%% retargeting method. We also analyze how varying the size of the data
%% -- e.g. the length of the edit, the lengths of the target video and
%% the amount of repository video -- impacts synthesis quality.
%
To evaluate our tool we analyze the quality of the synthesized video
as we vary the algorithmic methods (e.g. fast phoneme search, neural
retargeting) used in our synthesis pipeline, and as we vary the data
(e.g. length of target video, repository or edit) provided to the
algorithm.
We then compare the quality of our results to those of
previous work. Finally we report on user studies that quantitatively
evaluate the quality of our synthesized results.  Unless otherwise
noted, the evaluations in this section use results from our automatic
synthesis pipeline with no additional user refinement.  Readers should
refer to supplemental materials to evaluate the video results
presented in this section.
% \maneesh{Supplemental materials will need
  % to be updated with new evaluations.}

% \maneesh{We should think about all the evaluations we are planning to show and then organize them in some fashion. For example, we may first want to show that fast phoneme search produces results exactly the same as Fried 19. Then compare neural retargetting to alternative baselines, but also perhaps using the full target actor video -- e.g. Fried 19. I'm not sure.}

\subsection{Varying the Algorithmic Methods}
%\subsection{Evaluate Algorithms and Data Requirements}
\label{sec:eval_algo_data}

\paragraph{Comparison of phoneme search and stitching methods}
We compare the impact of using \new{ablated versions of} our fast phoneme search and
stitching algorithm (\Cref{sec:search}) with the phoneme search method
of Fried et al.~\shortcite{Fried_2019}.
More specifically, because their synthesis pipeline does
not include neural retargeting we treat Fried et al.'s approach as a baseline
method and build \new{two} comparison pipelines.
\new{The first one adds only phoneme {\em context expansion} (Section~\ref{sec:search}) to the baseline stitching method.
The second pipeline replaces their phoneme search and stitching method with the full version of our fast method (we call this version of the pipeline ``Modified Fried et al.~\shortcite{Fried_2019}'' in later comparisons).}
All three pipelines
assume access to an hour of target video which serves as the
repository.
%
%To compare the impact of using only our fast phoneme search and
%stitching algorithm on synthesis quality, we replace our neural
%retargeting method with the mthod of Fried et al. which assumes access
%to a large amount of target video (typically an hour).
%
%\maneesh{I don't follow the next snentence. What is the ``same hour-long target video''? How does our tool work with such a long target video? I think we need to explicitly say something like...  }
%With the same hour-long target video, we
%show result videos from Fried et al.\,\shortcite{Fried_2019}, as well as
%a version of Fried et al.\,\shortcite{Fried_2019} using our fast phoneme
%search and stitching algorithm.
\Cref{fig:eval_search} and videos in the supplemental materials show
that results generated using our fast phoneme search and stitching
method are often indistinguishable from the those generated by the
baseline. The main differences that do appear are often subtle as our
forced mouth closure on \textbackslash m, \textbackslash b, and
\textbackslash p phonemes reduces open mouth artifacts and our new
context aware stitching (Section~\ref{sec:search}) across subsequence transitions yields smoother,
less jittery lip motions.
%\maneesh{Is the jitter lip motion visible in
%  the figure and/or in the video comparison?}
%\david{jitter is not shown in the figure. It is visible in the video (at the end of ``people'' but is so subtle that we mostly focused on the mouth-non-closure at the first ``p'').}
More importantly our method takes only seconds to run, which is three
orders of magnitude faster than the hours required by baseline Fried
et al.~\shortcite{Fried_2019}, making it possible for the user to
iterate on edits. 

%% Compared to the baseline, results
%% using our fast method contain less jitter in mouth motions and can close the
%% mouth when the baseline fails to do so (\Cref{fig:eval_search}).
%% Furthermore, our algorithm takes only seconds to run, which is three
%% orders of magnitude faster than the hours required by baseline Fried
%% et al~\shortcite{Fried_2019}, making it possible for the user to
%% iterate on edits.

\begin{figure}
  \centering
  \includegraphics[width=\linewidth]{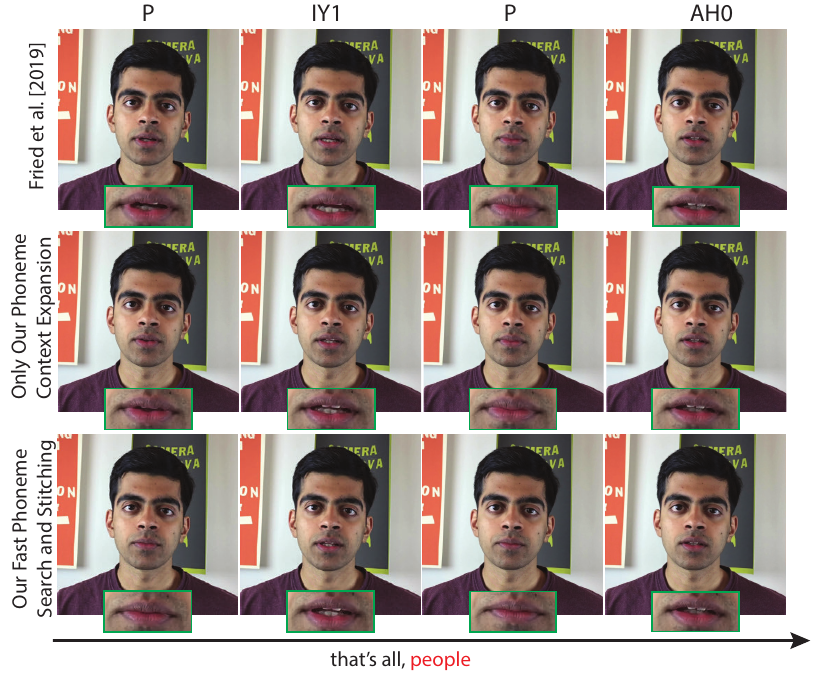}
  \vspace{-2em}
  \caption{Comparison of phoneme search and stitching method. \new{While the three methods usually yield visually indistinguishable frames, both the baseline method in Fried et al.~\shortcite{Fried_2019} and that with phoneme context expansion fail to close the mouth at the first ``P'' in people, whereas our fast phoneme search and stitching algorithm fully closes it.}}
  \label{fig:eval_search}
\end{figure}

  % \maneesh{This paragraph jumps around a bunch and I'm not really sure I understand what the main points are we are trying to make. It should not talk about the user study as this section is not about the user study. I don't undersand that it means to ``appear more real''. I don't think that is how we should describe our study results.}

\begin{figure}
  \centering
  \includegraphics[width=\linewidth]{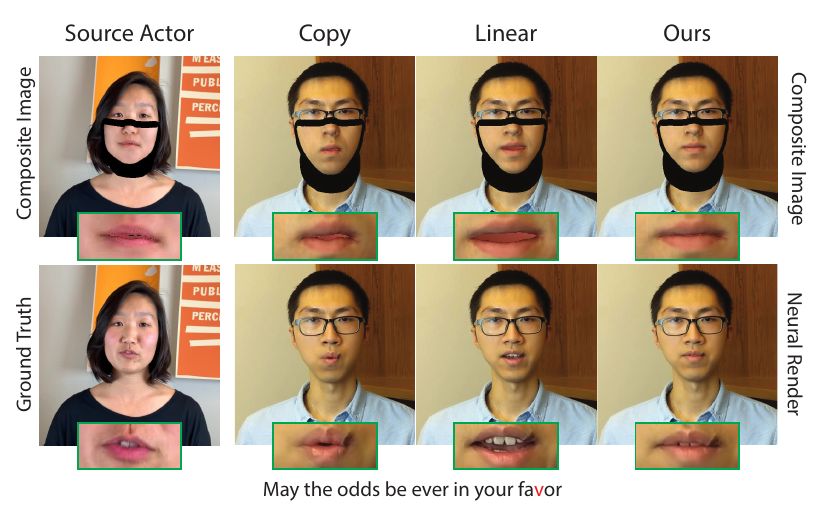}
  \vspace{-2em}
  \caption{Comparison of retargeting methods.
    Copying expression parameters yields an unnatural, rounded mouth shape on the target actor,
    while 
    linear regression fails to close his mouth for the ``v'' sound in the word ``favor''.
    In contrast, our retargeting method achieves a good match in lip shape to the source actor
    and properly matches the shape required for the ``v'' sound.}
  \label{fig:eval_retarget}
\end{figure}

\paragraph{Comparison of retargeting methods}
Our neural retargeting method (\Cref{sec:retargeting}) transforms a sequence of
source actor expression parameters to those of the target actor.  We
compare our method with two simpler baseline
retargeting methods.  The {\em copy} baseline directly copies the
source actor parameters to the target actor.  The {\em linear}
baseline replaces our retargeting network with a linear model.
%To build the linear model, we manually chose 2.5 minutes of target
%video containing phrases that exactly matched phrases in the source
%repository and then applied linear regression on the corresponding source and target
%parameter pairs.
More specifically, we manually chose 2.5 minutes of target video
containing phrases that exactly matched phrases in the source
repository and established a frame-to-frame correspondence by re-timing
the target phonemes to match the lengths of the source phonemes.  We
then applied linear regression on the source and target parameter
pairs to obtain the linear baseline model.
%To build the linear model, we manually established a frame-to-frame
% correspondence between phoneme subsequences in 2.5 minutes of the target video and
% those of the repository, by making sure that the target video
% contained the same phrases as the repository. We then applied linear regression
% on the source and target parameter pairs to obtain the model
% \maneesh{Can we tighten previous two sentences?.. Specifically, we chose 2-3 minutes of target video containing a phrases that exactly matched phrases in the source repository and then applied linear regression on the source and target parameter pairs.}
%\ohad{do we want to
%  emphasize that we don't do this for our method, only for the linear
%  baseline?}
% We then apply linear regression on the 
% source and target parameter pairs to obtain the linear retargeting
% model.
%we
%find a subset of the source repository with the same
%transcript as the target video, and use the phoneme boundaries to
%derive a frame-to-frame correspondence between the source and target
%actor.
%We then obtain the linear retargeting model by performing a
%linear regression on the source/target parameter pairs resulting from
%the frame-to-frame correspondence.  
\Cref{fig:eval_retarget} and the supplemental materials  show that our neural retargeting model
produces the best
results, while direct copying produces uncanny mouth shapes and the
linear model often fail to close the mouth, causing out-of-sync lip
motions.
%Full video comparisons are in supplemental material.
%\maneesh{Point to specific place in supplemental materials if possible.}

\subsection{Varying the Amount of Data}

\begin{figure}
\centering
\includegraphics[width=\linewidth]{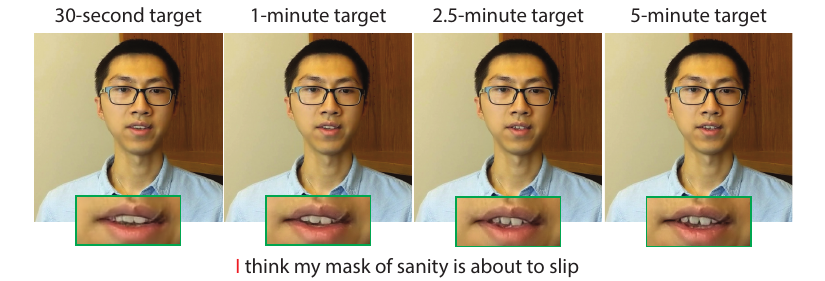}
\vspace{-1.5em}
\caption{With 2.5 minutes or more target video, our approach produces relatively sharp, realistic frames.
With only 30 seconds or 1 minute of target video, the frames become noticeably blurry especially around the mouth and teeth
(Zoom in at the front teeth to see the difference).
}
\label{fig:eval_target_size}
\end{figure}

\paragraph{Varying the length of the target video}
We examine the effect of different amounts of target video by applying
our tool on 10 minute, 5 minute, 2.5 minute, 1 minute
and 30 second subsets of a target video.  More target video
generally results in sharper images, higher quality mouth interiors
and smoother mouth motions. But the difference is subtle when target
video exceeds 5 minutes, and at 2.5 minutes the results remain
plausible.  With a 30 second target video however, although the
  results still have well-synchronized mouth motions, our neural
  renderer struggles and produces noticeably blurrier images, as shown in
  \Cref{fig:eval_target_size} and videos in supplemental materials.
Please zoom in on the front teeth to see the difference.
% Temporally, with limited target data our neural retargeting model
% produces choppier mouth motions, and the neural renderer produces less
% stable teeth that can appear to move with the lips.
% \maneesh{Not sure we need to say refer to supplemental material again if we said it up front. I think we should just refer to the supplemental material briefly when we introduce the figures -- see first two examples in section 5.1. Weird that Fig 9 does not include 10 min case, but maybe that is ok. Prev. version of the paper had a different mouth. shape. I can't see the differences really at all at the regular page size. The differences are kind of visible on zoom in on the teeth. We will need to ask reader to zoom in to see the differences, likely both here in the paragraph and in the caption.}
%\maneesh{Weird that Fig 9 does not include 10 min case, but maybe that is ok.}
% \maneesh{This section suggests that we need about 3 min of target video -- but up to now we've been suggesting 2-3 min is enough. Not sure what to make of that.}

\begin{figure}
  \centering
      \vspace{-1em}
  \includegraphics[width=\linewidth]{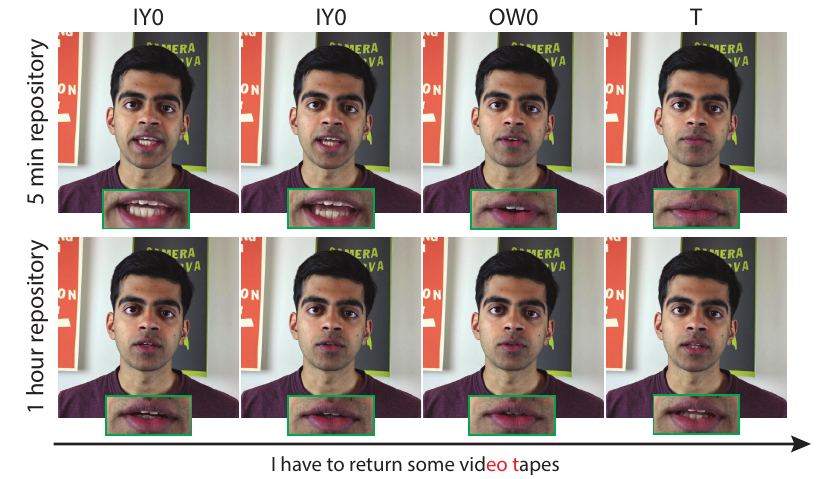}
  \vspace{-0.25in}
\caption{Effect of small repository size.
As shown here, the result from the 5-minute repository look less temporally stable as the transition from IY0 to OW0 is more drastic than the result from the full-hour repository.
With 5-minute repository, the actor also has an incorrect closed-mouth at T because we do not have as good a selection of phoneme coarticulations with a small repository as we do with a full-hour one.}
\label{fig:eval_repo_size}
\end{figure}

\begin{figure}
  \centering
    \vspace{-1em}
    \includegraphics[width=\linewidth]{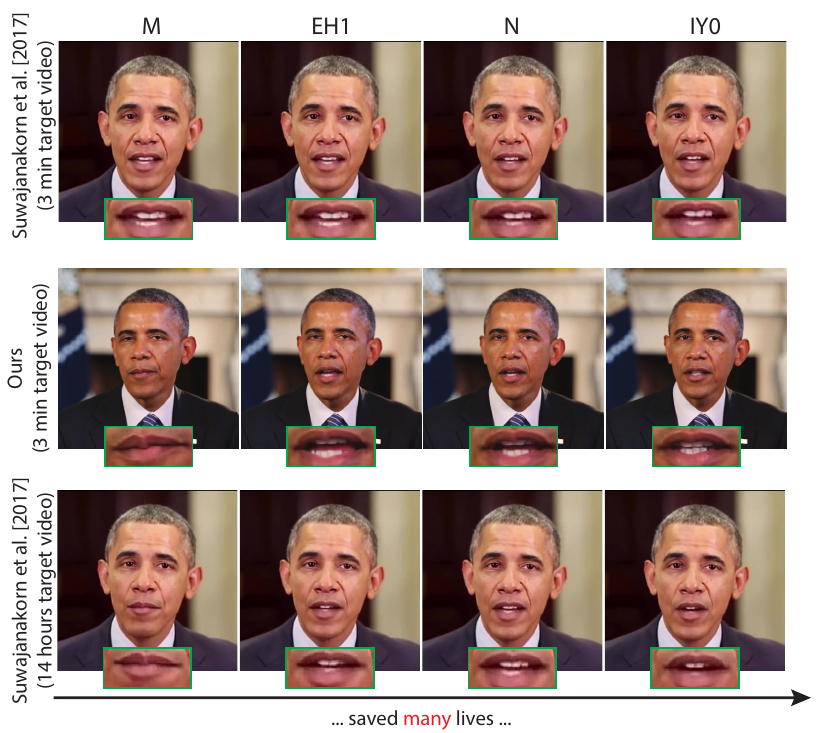}
      \vspace{-0.25in}
  \caption{Comparison to Synthesizing Obama\,\cite{10.1145/3072959.3073640}.
  With 3 minutes of Obama video, Suwajanakorn et al.\,\shortcite{10.1145/3072959.3073640} cannot give realistic mouth motions (top row).
  It produces similar frames throughout the word ``many'', and in particular fails to close the mouth at \textbackslash m.
  Our method produces well-synchronized mouth shapes and closures with only 3 minute of target video (middle row), while Suwajanakorn et al.\,\shortcite{10.1145/3072959.3073640} needs 14 hours of target video to produce a good result (bottom row).
  }
  \label{fig:eval_obama}
\end{figure}

% \begin{figure}
%   \centering
%       % \vspace{-1.5em}
% \includegraphics[width=\linewidth]{figs/edit_length.pdf}
% \caption{Effect of edit length. \maneesh{This figure appears mislabeled. Or perhaps I don't understand how it relates to edit length. Should label full sentence row. I guess one question here is what this has to do with length. If we were synthesizing popularity at the beginning of the sentence would we have the same issue? -- Perhaps this one should only be in the video -- we should discuss it.}
% \david{Your observation is correct. Here we pinpoint one artifact with our synthesis result, and basically say that as edits get longer these kinds of artifacts are more likely to emerge in greater numbers. It is not that the length of the edit itself is problematic, but that longer edits are more likely to contain phoneme sequences that get badly synthesized.}
% As shown here, when compared to the ground-truth, the actor in the full sentence synthesis does not open his mouth enough at ``a'' in popularity, leading to an unnatural artifact. As the edit length gets longer, such artifacts are more likely to emerge.
% }
% \label{fig:eval_edit_length}
% \end{figure}

\paragraph{Varying the amount of repository video}
We examine the effect of different amounts of repository video by
applying our tool on a 3.5 minute target video with 60 minute, 30
minute, 10 minute and 5 minute subsets of repository data.  Generally
a larger repository leads to better results. 
\Cref{fig:eval_repo_size} and videos in supplemental materials show
that as the repository shrinks, mouth motions become choppier as it becomes harder to find
long matching phoneme subsequences in the repository and our tool has
to introduce more transitions. However, the quality degrades
gracefully.
% \maneesh{Point to figure and supplemental material earlier in paragraph using the same approach as used in the first two cases in  Section 5.1}

\paragraph{Varying the length of the edit}
We examine the effect of synthesizing edits of different lengths by
synthesizing increasingly longer portions of the sentence ``only the most accomplished artists obtain popularity''; starting by only synthesizing the first word and sequentially
adding words until the full sentence is synthesized by our pipeline.
Videos in supplemental materials show that,
while the full sentence synthesis still has good mouth
motions and looks plausible, it generally produce results
that contain more artifacts than shorter edits, since with
more phonemes to synthesize there are more opportunities for artifacts to emerge.
% \maneesh{What is the sentence. Need to point to a figure here as well as supplemental videos (see how we wrote it in earlier sections).}
% Compared to the ground-truth
% recording, while the full sentence synthesis still has good mouth
% motions and looks plausible, shorter edits generally produce results
% that contain smoother mouth motions and fewer artifacts, since with
% more phonemes to synthesize there are more opportunities for artifacts to emerge.
% \maneesh{We should try to say why these artifacts appear. Why are there more artifacts as the synthesis gets longer.}
% \maneesh{Should probably point to user study if we keep this paragraph in the main paper.}
Our user studies (\Cref{sec:user_study}) also found that results from short edits are
rated more real than full-sentence syntheses.
% \david{May get moved to supplementary PDF.} \maneesh{this one probably needs to stay and we should try to find an example that fails that we can make a figure of. I think it should stay because the difference between short and long edits is a big deal in our user study.}

\begin{figure}
  \centering
      \vspace{-1em}
      \includegraphics[width=\linewidth]{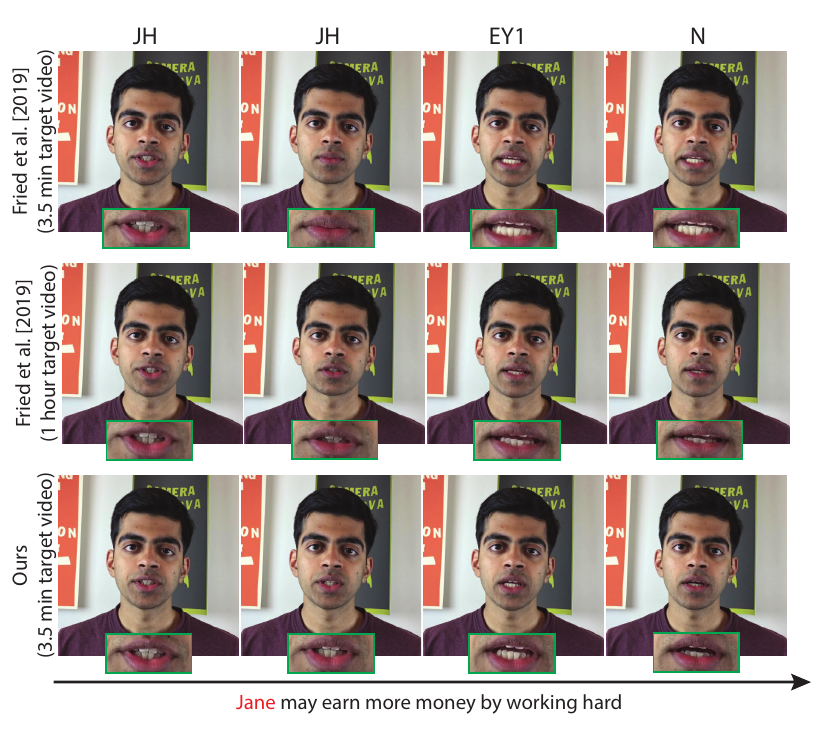}
          \vspace{-0.25in}
  \caption{Comparison to Fried et al.\,\shortcite{Fried_2019}.
  Mouth motions from Fried et al.\,\shortcite{Fried_2019} are less temporally coherent as shown from phoneme 'JH' to 'EY1',
  both with 3.5 minutes (top row) and with 1 hour (middle row) of target video.
  With 3.5 minutes, Fried et al.\,\shortcite{Fried_2019} also produced an incorrect mouth closure during the second 'JH' frame.
  Our results (bottom row) have smoother mouth motions, and we use only 3.5 minutes of target video.
  }
  \label{fig:eval_fried}
\end{figure}

\begin{figure}
  \centering
        \vspace{-1em}
  \includegraphics[width=\linewidth]{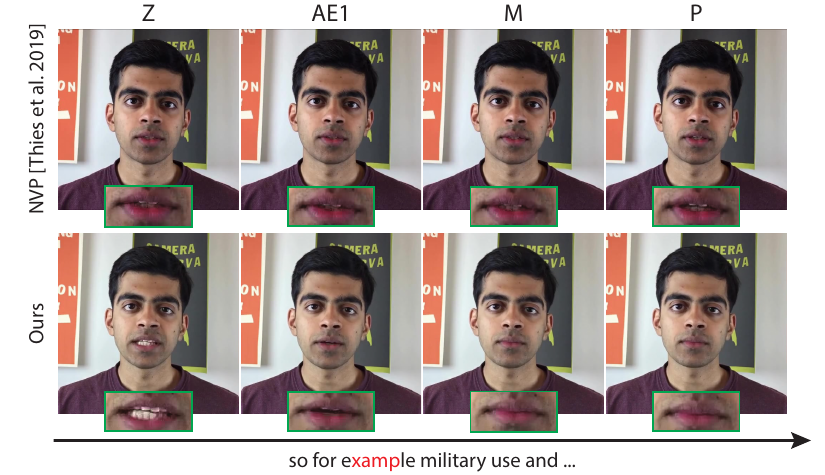}
            \vspace{-0.25in}
  \caption{Comparison to Neural Voice Puppetry (NVP)\,\cite{thies2019neural}. The actor fails to close his mouth with Neural Voice Puppetry at \textbackslash m and \textbackslash p, whereas our approach with no user refinement fully closes it. We refer to the supplemental materials for video comparison.
  % \maneesh{Need to point out specific errors.}
  }
  \label{fig:eval_nvp}
\end{figure}

\begin{figure}
  \centering
  \vspace{-1em}
  \includegraphics[width=\linewidth]{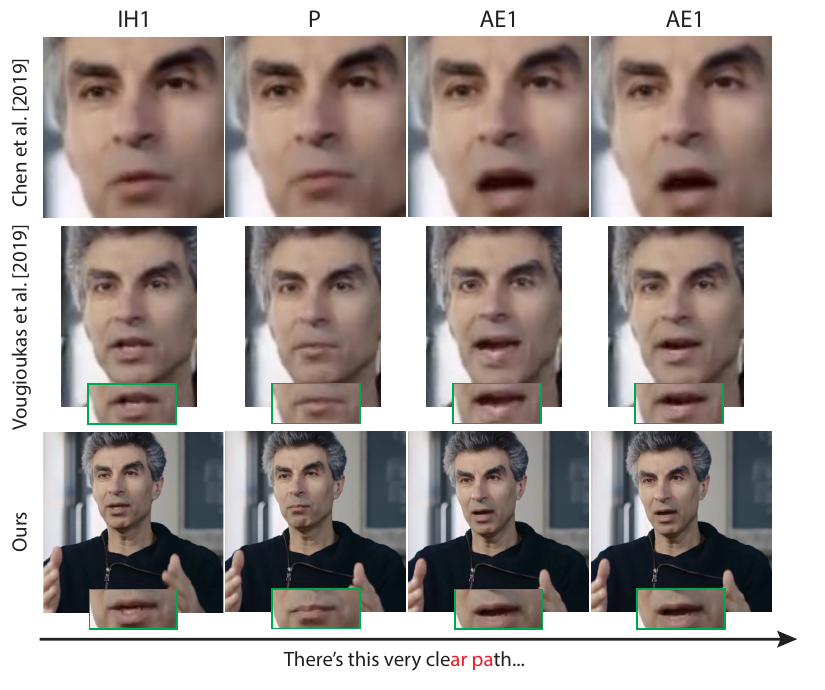}
  \vspace{-0.25in}
  \caption{Comparison to Chen et al.~\shortcite{chen2019hierarchical} and Vougioukas et al.~\shortcite{Vougioukas2019}.
  \new{All three methods produce good lip-sync, but our result has better quality due to sharper images and natural head motion.}
  }
  \label{fig:eval_c19_v19}
\end{figure}

\subsection{Comparisons to Other Methods}
\label{subsec:compare_others}

\paragraph{Comparisons with Synthesizing Obama\,\cite{10.1145/3072959.3073640}}
Suwajanakorn et al.\,\shortcite{10.1145/3072959.3073640} have
presented a technique for taking audio speech of Obama as input and
synthesizing a video of Obama saying the speech. We compare our
results to theirs using only 3 minutes of video of Obama.
As shown in \Cref{fig:eval_obama}, our approach
gives more plausible and synchronized mouth motions compared to their
method which can fail to close the mouth. While our approach produces good results
with only 3 minutes of Obama video, their method needs 14 hours.
 % \maneesh{What is more plausible. Point to the artifacts in
 %  theirs. Also why is the synthesized video not the same view of
 %  Obama. That would be a better comparison. Need to talk about 14hr Obama result -- what is the takeaway?}

\paragraph{Comparisons with Fried et al.~\shortcite{Fried_2019}}
Our work builds on Fried et al.'s~\shortcite{Fried_2019} synthesis
pipeline.  However, they require over an hour of target video to
produce high-quality results.  We compare our method against theirs
using 3 minutes of target video, as well as theirs using an hour of
target video.  \Cref{fig:eval_fried} and videos in supplemental
materials show that our result has more temporally coherent mouth
motions than Fried et al.~\shortcite{Fried_2019}, both when they use
the same 3 minutes of target video, and when they use 1 hour of target
video.  We further compare our results to those for Fried et
  al.\,\shortcite{Fried_2019} in the user studies in
  \Cref{sec:user_study}.

  \paragraph{Comparisons with Neural Voice Puppetry\,\cite{thies2019neural}}
Concurrent to our work, Neural Voice Puppetry (NVP) can synthesize
talking head videos from audio signal input, or from text by using a
synthetic voice to obtain the audio signal.  Given an audio speech
track, we compare our result to theirs by applying our method on
the phoneme timings extracted from the audio.
% \maneesh{use same convention as earlier for referring to Figure and supplemental materials. We should put all the comparisons in the supplemental.}
\Cref{fig:eval_nvp} and videos in supplemental materials show that
while both approaches generate mouth motions that synchronize with the
audio, our fully automatic result (without user refinement) generates
closed-mouth frames at the desired phonemes (\textbackslash m,
\textbackslash b, \textbackslash p), whereas Neural Voice Puppetry
leaves the mouth open for many of these phonemes.
In addition, unlike NVP, our tool allows the user to iteratively refine the automatic results and adjust the performance. We further compare our results to NVP in the user studies in \Cref{sec:user_study}.
% \maneesh{Should point to user study comparison.}
% \maneesh{Probably should remind reader that they don't provide any performance and refinement controls.}

\paragraph{Comparisons with Chen et al.~\shortcite{chen2019hierarchical} and Vougioukas et al.~\shortcite{Vougioukas2019}}
\new{
Both Chen et al.~\shortcite{chen2019hierarchical} and Vougioukas et al.~\shortcite{Vougioukas2019} generate talking head videos
from audio input and a single frame of the target actor.
Given an audio speech track, we compare our result to theirs by applying our method on the phoneme timings extracted from the audio.
\Cref{fig:eval_c19_v19} and videos in supplemental materials show that
while all three approaches produce good lip-sync with proper mouth closures,
both Chen et al.~\shortcite{chen2019hierarchical} and Vougioukas et al.~\shortcite{Vougioukas2019} produce videos of less resolution than our result.
In addition, their results do not have the natural head motion in our result, and by always centering the video around the cropped head, their results can contain warping artifacts in the background,
making them ill-suited for incorporation into a video-editing workflow.
}
% \david{Add paragraph about the full ablation study:
% Fried19 on full data, ours without retargeting on full data, 
% ours with retargeting but using GAN trained on full data,
% ours.} \maneesh{Let's write this out. I don't have a good sense of what all we are doing in the ablation.}
%
%% Notably, Fried~\etal~\shortcite{Fried_2019}
%% takes hours to generate a sentence with an hour of target video, and
%% still takes over 10 minutes on 3 minutes of target video, while our
%% method takes 40 seconds to produce the result.

\input{evaluation-userstudy}

%% file: evaluation-userstudy.tex
\subsection{User Studies and Automatic Metrics}\label{sec:user_study}

\new{We use both user studies and automatic metrics} to quantitatively evaluate the quality
of the video generated by our editing tool.
In the user studies, we investigate both short and long edits, while
ablating the target video length and the neural retargeting step. We
compare to previous work\,\cite{Fried_2019,thies2019neural} and to
ground-truth video.
We follow the study design used in previous talking-head synthesis
research\,\cite{kim2018DeepVideo,Fried_2019}.  Specifically,
participants see one video at a time in randomized order and are asked
to rate the statement ``This video clip looks real to me'' on a
5-point Likert scale ranging from strongly disagree (1) to strongly
agree (5).
% We use real human voice (recorded separately) so that participants can compare them with ground truth recordings. 
All videos used in our studies are available in the supplemental material.
% No subject participated in both studies.

\paragraph{User study 1: Short Phrases (1 -- 4 words)}
Short phrases are the main type of result shown in Fried et al.\,\shortcite{Fried_2019}. Such edits are useful for minor fix-ups on existing sentences. 
In this study we compare our automatic synthesis results (``Ours'') to 3
versions of Fried et al.\,\shortcite{Fried_2019}.
(1) Their method with the same amount of target video as used by our tool, which is less than 5 minutes in all
cases. 
(2) Their method with 1 hour of target video, which is their
recommended amount, and more than 12 times the amount we use.
(3) A version of their method with our fast
phoneme search and stitching algorithm (\Cref{sec:search}), but with 1
hour of target video (``Modified Fried et al.~\shortcite{Fried_2019}''
in \Cref{sec:eval_algo_data}), to evaluate the effect of ablating our
neural retargeting step. 
We also compare to ground truth video recordings.
We recruited 110 participants to view 25 videos each (5 conditions for each of 5 edits).
We report Likert scale responses in \Cref{tab:user_study} (``Short Phrase'').
The differences between all pairs, except ``Ours'' vs. ``Modified Fried'', are statistically significant. 
\new{All p-values have been adjusted for multiple testing and are reported in supplemental materials.}
% The difference between conditions is statistically significant (Kruskal-Wallis test, $p < 10^{-20}$).
%Among the synthesized videos, our tool and Modified Fried produce the highest-quality videos.  In particular,

Our tool outperforms Fried et al.\,\shortcite{Fried_2019} both when
using 5 minutes of data and 1 hour of data.  We believe this is due to
our results having more accurate mouth closures and better temporal
coherency in mouth motions.
Results are similar for our tool and
Modified Fried, indicating that our neural retargeting step does not
have much negative effect on result quality.
Together these results also suggest that our fast phoneme search with stitching that forces closed mouths for \textbackslash m, \textbackslash b, and \textbackslash p phonemes leads to higher-quality synthesis than the slow phoneme search and stitching approach used originally by Fried et al.\,\shortcite{Fried_2019}.
Although a gap still
remains between our synthesized results and ground-truth videos, our
results for short edits are rated as real almost two thirds of the time.
% Synthesis time with our fast phoneme search and stitching algorithm is orders of magnitude faster compared to Fried et al.~\shortcite{Fried_2019} (each result is synthesized in 40 seconds using our pipeline  vs. hours for their unmodified pipeline).

\paragraph{User study 2: Full Sentences (6 -- 9 words)}
Full sentence synthesis is more challenging, since longer synthesis equates to a larger chance of inaccurate matches and synthesis artifacts. However, synthesizing full sentences as opposed to short phrases opens up more use cases (\Cref{sec:results}). Investigating full-sentence synthesis also emphasizes the quality differences between methods.
The conditions in this user study are the same as for user study 1.
We recruited 153 participants to view 25 videos each (5 conditions for
each of 5 sentences).
We report Likert scale responses in \Cref{tab:user_study} (``Full Sentence'').
% The difference between conditions is statistically significant (Kruskal-Wallis test, $p < 10^{-72}$).
The differences between all pairs, except ``Fried < 5 min'' vs. ``Fried > 1 hr'', are statistically significant. 
\new{All p-values have been adjusted for multiple testing and are reported in supplemental materials.}

Our tool produces the highest-quality results, followed by Modified
Fried with over 1 hour of data, then by Fried et al.~\shortcite{Fried_2019}.  
Similar to user study 1, here our results
have better mouth closures and smoother mouth motions than Fried et
al.~\shortcite{Fried_2019}.  It is worth noting that our results are
even better than Modified Fried.  We believe this is because our tool has a higher-quality source repository which becomes more salient when the edits are long.  It shows the
advantage of our approach to decouple source repository from target
video, as data quality improvements to the repository can benefit many
different target videos. The one-time cost of building a high-quality repository amortizes across all the edits that use it.

% A followup Turkey's range test
% shows that our automatic results are higher-quality than Fried et
% al.~\shortcite{Fried_2019}, both when using the same amount of data
% ($p=0.001$) and when using 12x the amount of data ($p=0.001$).  Fried
% et al.~\shortcite{Fried_2019} with our fast phoneme search and
% stitching algorithm gives higher-quality results than unmodified Fried
% et al.~\shortcite{Fried_2019} ($p=0.001$), but our tool using 12x less
% data produced even higher quality results ($p=0.006$).  We believe
% this is because the source repository in our tool has higher-quality
% phoneme alignments as we have manually fixed errors from the automatic
% aligner, and longer edits are more likely to pick phonemes with bad
% alignments. This shows the advantage to decouple source repository
% from target video.
% \maneesh{See note on previous paragraph.} \maneesh{I'm not sure we should bring up manual fixes to the repository. I think we should attribute the difference to the better stitching and perhaps to a high-quality repository. I wouldn't focus on misalignments as being the issue but rather that our repository is large and perhaps that the source actor emphasizes mouth movements more than the other actors. Let's discuss this.}

\paragraph{User study 3: our tool vs Neural Voice Puppetry~\cite{thies2019neural}}
The third user study compares our results to those of Neural Voice
Puppetry\,\cite{thies2019neural}, where we show viewers videos generated by the
two methods from the same audio speech track. 
We recruited $90$ participants to view 8 videos each (4 from each of
the two methods, \Cref{tab:user_study3}).  The audio tracks used in
the videos are not the actor's real voice, and we believe this is the predominant reason for overall lower scores (for both methods). 
The difference between conditions is not statistically significant, and our results
have similar mean scores to those of NVP.
Nevertheless, as mentioned in \Cref{subsec:compare_others}, closely examining the videos generated by the two approaches, we find that our method often does a better job of closing the mouth on \textbackslash m, \textbackslash b, and \textbackslash p phonemes.
We also note that while our user studies evaluate our \emph{automatic} results, unlike NVP, our tool also provides refinement and performance controls that can be used to improve results over the course of an interactive editing session.

\paragraph{Automatic Metrics}
\new{
	For videos in user study 1 and 2 where we have ground-truth recordings, we further evaluate the results using automatic metrics against grouth truth videos.
	To measure reconstruction quality, we compute structural similarity index (SSIM \cite{SSIM2004}) and peak signal-to-noise ratio (PSNR).
	To measure lip motion quality, we compute the Landmarks Distances (LMD) \cite{chen2018lip}.
	We report the results in \Cref{tab:autometrics}.
	Although our methods achieve favorable scores on many of these metrics, the score differences are quite small and visual differences can be subtle.
	We believe the user studies provide a better and more trustworthy measure of video quality.
}

\begin{table}
\footnotesize
\centering
\begin{tabular}{@{}lllccccccc@{}}
\toprule
 &         &       & \multicolumn{5}{c}{Likert response (\%)} & & \\
      \cmidrule(lr){4-8}    
&Condition & Length & 5 & 4 & 3 & 2 & 1 & Mean & `Real' \\
%Condition                           & Length of    & 5 (Strongly           & 4      & 3      & 2      & 1 (Strongly              & Mean         & \% `Real'       \\
%                                    & target video & \hspace{0.1cm}agree)  &        &        &        & \hspace{0.44cm}disagree) & score        &                 \\
\midrule
\parbox[t]{0mm}{\multirow{5}{*}{\rotatebox[origin=c]{90}{\footnotesize Short Phrase}}}
&Fried et al.~\shortcite{Fried_2019}        & < 5 min      & 19.5 & 28.0 & 11.3 & 22.1 & 19.0 		& 3.1 					& 47.6\%          \\
&Fried et al.~\shortcite{Fried_2019}       & > 1 hr       & 24.1 & 31.5 & 13.7 & 20.0 & 10.7 		   	& 3.4 		  			& 55.6\%          \\
&Modified Fried		& > 1 hr      & 27.5 & 39.9 & 13.3 & 13.9 & 5.4 			& \textbf{3.7} 			& \textbf{67.4\%} 			\\
&Ours			    & < 5 min      & 30.37& 34.2 & 14.1 & 15.2 & 5.7 			& \textbf{3.7}  		& \textbf{64.9\%}		 \\[2px]
&Ground truth        & n/a          & 40.1 & 37.8 & 11.5 & 9.8 & 0.9 			& 4.1 					& 77.8\%          \\
\midrule
\parbox[t]{0mm}{\multirow{5}{*}{\rotatebox[origin=c]{90}{\footnotesize Full Sentence}}}
&Fried et al.~\shortcite{Fried_2019}       & < 5 min      & 14.7 & 22.9 & 11.1 & 25.0 & 26.3 		& 2.7 					& 37.6\%         \\
&Fried et al.~\shortcite{Fried_2019}       & > 1 hr       & 14.6 & 22.5 & 12.2 & 26.3 & 24.5 		& 2.8 					& 37.1\%          \\
&Modified Fried		& > 1 hr      & 16.5 & 31.5 & 14.1 & 20.7 & 17.2 		& \textbf{3.1} 			& \textbf{48.0\%}			\\
&Ours			    & < 5 min      & 17.9 & 38.2 & 16.0 & 20.1 & 7.7 		& \textbf{3.4} 			& \textbf{56.2\%}		 \\[2px]
&Ground truth        & n/a          & 39.0 & 42.1 & 7.3 & 8.8 & 2.9 			& 4.1 					& 81.1\%          \\
\bottomrule
\end{tabular}
\caption{Results from user studies on short phrases and full sentences.
The ``Length'' column shows the length of the input target video for each method.
% \maneesh{``Length'' heading should be ``Repo Size'' or some such. Fried et al. should be ``Fried et al. [2019]''}
%   \ohad{adding [2019] requires changing to a 2 column layout with a lot of wasted whitespace---I don't think it's worth it. Is ``Repo Size'' accurate here? Perhaps David can chime in.} \maneesh{I added in [2019] to the first row. Seems to fit. What is the issue?}
% \david{``Repo Size'' is incorrect. This should be ``Target Video Length'' but that is too long, so we shortened it to ``Length'' in the previous submission.} \maneesh{Ok, but then caption needs to explain up front what length means. -- it is the length of the input target video.}
We compare to previous work \cite{Fried_2019} and to ground-truth recordings.
We report percentage of each answer on a 5-Point Likert scale, the mean score, and percent of videos that received a score of 4 or 5 (`real').
The difference between conditions is significant in both studies (Kruskal-Wallis test, $p < 10^{-20}$ each).
A followup Tukey's range test shows that 
all pairwise comparisons are statistically significant ($p < 0.008$ each) except for
``Ours'' vs. ``Modified Fried'' for short edits and
``Fried < 5 min'' vs. ``Fried > 1 hr'' for full sentences.
\new{
	Note that Tukey's procedure adjusts the p-values for multiple comparisons. We report all adjusted p-values in the supplemental materials.
}
Using our fast phoneme search and stitching algorithm improves results from Fried et al.~\shortcite{Fried_2019}.
Our tool outperforms the method of Fried et al.~\shortcite{Fried_2019} when they use the same amount of target video data,
and when they use 12x the amount of data.
For full-sentence synthesis, our tool also outperforms Modified Fried even while they use 12x the amount of data. 
}
\label{tab:user_study}
\end{table}

\begin{table}
\footnotesize
\centering
\begin{tabular}{@{}llccccccc@{}}
\toprule
         &       & \multicolumn{5}{c}{Likert response (\%)} & & \\
      \cmidrule(lr){3-7}    
Condition & Length & 5 & 4 & 3 & 2 & 1 & Mean & `Real' \\
%Condition                           & Length of    & 5 (Strongly           & 4      & 3      & 2      & 1 (Strongly              & Mean         & \% `Real'       \\
%                                    & target video & \hspace{0.1cm}agree)  &        &        &        & \hspace{0.44cm}disagree) & score        &                 \\
\midrule
Ours 		& < 5 min 	& 6.0 & 9.6  & 10.6 & 26.2 & 47.5 		& 2.0	& 15.6\%		\\
NVP			& < 5 min   & 6.0 & 10.6 & 11.3 & 24.8 & 47.4		& 2.0	& 16.6\%		\\ 
\bottomrule
\end{tabular}
\caption{Results from user study on our tool and Neural Voice Puppetry (NVP)~\cite{thies2019neural}.
We compare to NVP and report percentage of each answer on a 5-Point Likert scale, as well as mean score and percent of videos that received a score of 4 or 5 (`real').
Our tool and NVP received similar mean scores and the difference is not statistically significant (Kruskal-Wallis test, $p = 0.84$).
}
\label{tab:user_study3}
\end{table}

\begin{table}
\small
\centering
\begin{tabular}{@{}lllccc@{}}
\toprule
&Condition & Length & SSIM & PSNR & LMD \\
\midrule
\parbox[t]{0mm}{\multirow{4}{*}{\rotatebox[origin=c]{90}{\footnotesize Short Phrase}}}
&Fried et al.~\shortcite{Fried_2019}       & < 5 min & 0.89929 			& 25.08146 			& 4.57492 \\
&Fried et al.~\shortcite{Fried_2019}       & > 1 hr  & 0.89925 			& 25.09057 			& 4.47108 \\
&Modified Fried							   & > 1 hr  & \textbf{0.89934} & \textbf{25.09670} & \textbf{4.38139} \\
&Ours						   			   & < 5 min & 0.89921          & 25.09349          & 4.57060 \\
\midrule
\parbox[t]{0mm}{\multirow{4}{*}{\rotatebox[origin=c]{90}{\footnotesize Full Sentence}}}
&Fried et al.~\shortcite{Fried_2019}       & < 5 min & 0.96339			& 32.74630 			& 3.75896 \\
&Fried et al.~\shortcite{Fried_2019}       & > 1 hr  & 0.96552          & 32.94852			& 3.67367 \\
&Modified Fried							   & > 1 hr  & 0.97578          & 35.00073			& \textbf{2.90959} \\
&Ours			    					   & < 5 min & \textbf{0.97630} & \textbf{35.12082} & 3.18670 \\
\bottomrule
\end{tabular}
\caption{
\new{Results of automatic metrics.
We compare our results to 3 versions of Fried et al.~\shortcite{Fried_2019} for both short phrase edits and full sentence syntheses.
Best score is bolded. Our tool tops SSIM and PSNR for full sentence syntheses and ranks second after Modified Fried for LMD on full sentence and PSNR on short edits.
}}
\label{tab:autometrics}
\end{table}

%% file: futureWork.tex
\section{Limitations and Future Work}

%% Our work is motivated by the

%% We motivated our work with the need for an iterative, interactive editing tools 

%% A practical tool for editing talking-head video must support an interactive/iterative editing workflow and

%% An interactive editing loop is essential for many real-world video editing 

We have demonstrated an iterative text-based
tool for editing talking-head dialogue and performance that can be
applied to many real-world editing scenarios in which only a few
minutes of target actor video is available. However, our approach does
have several limitations that could be addressed in future work.

\paragraph{Further reduce feedback loop time}
Our tool currently requires about $30$ seconds to synthesize a
typical $5$ word edit. While this feedback loop time allows users to
try a variety of edits and refinements, seeing a synthesized result
immediately (in real-time) would allow even more iteration and
exploration of design space. As noted in Section~\ref{sec:runtime},
parallelization of our fast phoneme search and neural rendering steps
as well as streaming playback of the synthesized video could reduce
the feedback loop time significantly.

\paragraph{Improve quality of synthesis results}
Although our method compares favorably in quality with previous
talking-head synthesis techniques (\Cref{sec:user_study}), there is
still a gap in realism between our results and ground-truth videos.
As our method relies on a rich repository of source video to provide
mouth motions for phoneme coarticulations, it may be possible to
improve synthesis by developing higher-quality repositories.
One approach may be to leverage existing work in text-driven 3D human
mouth animation\,\cite{edwards2016jali} to render unlimited amounts of
mouth motions to serve as the repository.
Another direction is to build multiple repositories of many different
source actors and then given a target video, develop techniques to
pick the best source actor for the target.

\paragraph{Performance controls over full face}
Our current approach focuses on synthesizing lip motions that match
the target edit. While our tool offers controls for inserting mouth
gestures and changing the speaking style, the effects of these
controls are limited to the lower part of the face.
%and we rely on a
%retimed segment of the original target video to provide the
%rest of the head.
Others have demonstrated techniques for controlling more of the head,
including the ability to change head pose, gaze direction and whole
face expressions\,\cite{kim2018DeepVideo} However, these techniques
often introduce artifacts in the hair and with the clothes at the
neckline.  Adding such full face controls in an artifact-free manner
remains an open research direction.

\paragraph{Previsualizing dialogue using existing film scenes}
When writing dialogue, scriptwriters have to imagine the sound
and appearance of the scene. Using our video editing tool with a
catalog of video from existing film scenes might allow such
scriptwriters to quickly visualize the dialogue in different settings
and with different actors. Users might search for scenes based on
their settings and actors using a tool like
SceneSkim\,\cite{pavel2015sceneskim} and our tool could be used to insert
the new dialogue.

%For example, users could see how their dialogue works
%when Clint Eastwood speaks it as a cowboy in a Western, or when Samuel
%Jackson speaks in a superhero film. \maneesh{ok I'm not sure about
%  this one.} \ohad{I would remove the entire paragraph}

% LocalWords:  iteratively retimed Previsualizing SceneSkim
% LocalWords:  coarticulations

%% file: ethics.tex
\section{Ethical Considerations}
%\todo{Acknowledge potential for misuse, emphasize legitimate use cases and reiterate ethical principles (consent and transparency) of such technology.}

%% Our tool for editing talking-head video is designed to facilitate an
%% iterative workflow common in many video editing tools today.
%% Whether an editor trying different way to phrase the dialogue in a
%% film, developing dialogue for a conversational agent, or correcting a
%% mistake in a lecture, such iteration is often essential for finding
%% the most appropriate result.
%% %Such iteration is often essential for removing filler words, adjusting
%% %the phrasing and correcting mistakes in video.
%% \maneesh{May not need second sentence. I think we should try to keep this short.}

Our editing tool is designed to enable an iterative workflow for
removing filler words, adjusting phrasing, or correcting mistakes in a
talking-head video. While such tools can facilitate content creation and storytelling,
tools like ours, that let users manipulate what a target actor is
saying, can also be misused.
%However, editing tools like ours, that let users manipulate what a
%target actor is saying, can be misused.
We follow the guidelines suggested by Fried et
al.\,\shortcite{Fried_2019} for ethically using such tools.  (1) Video
generated by our tool should be {\em transparent} about the fact that
it has been manipulated.
%\ohad{consider removing the next sentence}
%Such transparency may be clear from context (e.g. it is clearly parody) or may be signaled
%explicitly (e.g. via watermarking). 
(2) Actors must give {\em consent}
to any manipulation before a resulting video is
shared widely.

We also recognize that these guidelines alone will not stop bad actors
from using tools like ours to create false statements and slander
others. Therefore, it is also critical for researchers to continue
developing
%\ohad{alternative: "Therefore, we are also developing ..."
%  (true statement since I'm working with Shruti on this)}
tools for
detecting, fingerprinting, and verifying such video
manipulation. Openly publishing the technical details of our tool
%manipulation tools 
can increase public awareness and help 
%the researchers' work on 
detection efforts. 
Ultimately these issues may also require regulations and laws that
penalize misuse while allowing creative and consensual use cases.

%\maneesh{Maybe cite our work and others on detecting manipulated video?}
%\david{Agarwal et al.~\shortcite{agarwal2020} is already cited in the method section, but no objection citing it and others here.}

%\maneesh{May be too much of a copy of Fried et al. - especially last
%paragraph.}

%\ohad{maybe place this section before the conclusion, so that we end
%on a positive note}

% LocalWords:  et al

%% file: conclusion.tex
\section{Conclusion}

Iterative editing is central to many content-creation tasks and is
especially common in video editing.  We have shown how to enable such
iterative editing in the context of editing talking-head video using a
text-based interface that allows changes to wording and facial
performance while providing refinement controls.  Whether an editor
trying different ways to phrase the dialogue in a film, developing
dialogue for a conversational agent, or correcting a mistake in a
lecture, such iteration is often essential for finding the most
appropriate result. We believe such tools that facilitate video
editing can democratize content-creation and enable many more people
to tell their stories.

%Such iteration is often essential for removing filler words, adjusting
%% %the phrasing and correcting mistakes in video.
%% \maneesh{May not need second sentence. I think we should try to keep this short.}